%% file: arxiv_main.tex
\PassOptionsToPackage{svgnames}{xcolor}
\documentclass{article}

\usepackage{arxiv}

\usepackage[utf8]{inputenc} 
\usepackage[T1]{fontenc}    
\usepackage{hyperref}       
\usepackage{url}            
\usepackage{amsmath}
\usepackage{amssymb}
\usepackage{booktabs}
\usepackage{amsfonts}       
\usepackage{nicefrac}       
\usepackage{microtype}      
\usepackage{natbib}

\usepackage{txfonts}
\usepackage{graphicx}
\usepackage{csquotes}
\usepackage{tikz}
\usepackage{wrapfig}
\usepackage{cleveref}
\usepackage{multirow}
\usepackage{pifont}
\usepackage[many]{tcolorbox}

\definecolor{main}{HTML}{5989cf}    
\definecolor{sub}{HTML}{cde4ff}     
\definecolor{spurious_purple}{HTML}{CC00FF}

\newtcolorbox{boxD}{
    colback = sub, 
    colframe = main, 
    boxrule = 0pt, 
    toprule = 3pt, 
    bottomrule = 3pt, 
    title = Running Example: Classifying Birds into Landbirds and Waterbirds,
}

\newcommand{\cmark}{\textcolor{ForestGreen}{\ding{51}}}%
\newcommand{\xmark}{\textcolor{DarkRed}{\ding{55}}}%

\setcitestyle{numbers,open={[},close={]}}

\title{Navigating Shortcuts, Spurious Correlations, and Confounders: From Origins via Detection to Mitigation}

\author{
\normalfont
\textbf{David Steinmann}\textsuperscript{1,2} \qquad \textbf{Felix Divo}\textsuperscript{1} \qquad \textbf{Maurice Kraus}\textsuperscript{1} \qquad \textbf{Antonia Wüst}\textsuperscript{1}\vspace{0.5ex}\\
\textbf{Lukas Struppek}\textsuperscript{1,3} \qquad \textbf{Felix Friedrich}\textsuperscript{1,2} \qquad \textbf{Kristian Kersting}\textsuperscript{1,2,3,4}\vspace{1.5ex}\\
\textsuperscript{1}AI \& ML Group, TU Darmstadt \quad \textsuperscript{2}Hessian Center for AI (hessian.AI) \\ \textsuperscript{3}German Research Center for AI (DFKI) \quad \textsuperscript{4}Centre for Cognitive Science, TU Darmstadt \vspace{0.5ex}\\
\texttt{david.steinmann@tu-darmstadt.de}\\
}

\begin{document}
\maketitle

\begin{abstract}
Shortcuts, also described as Clever Hans behavior, spurious correlations, or confounders, present a significant challenge in machine learning and AI, critically affecting model generalization and robustness. Research in this area, however, remains fragmented across various terminologies, hindering the progress of the field as a whole. Consequently, we introduce a unifying taxonomy of shortcut learning by providing a formal definition of shortcuts and bridging the diverse terms used in the literature. In doing so, we further establish important connections between shortcuts and related fields, including bias, causality, and security, where parallels exist but are rarely discussed. Our taxonomy organizes existing approaches for shortcut detection and mitigation, providing a comprehensive overview of the current state of the field and revealing underexplored areas and open challenges. Moreover, we compile and classify datasets tailored to study shortcut learning. Altogether, this work provides a holistic perspective to deepen understanding and drive the development of more effective strategies for addressing shortcuts in machine learning.
\end{abstract}


\input{body}

\section*{Acknowledgements}
This work was supported by the Priority Program (SPP) 2422 in the subproject \enquote{Optimization of active surface design of high-speed progressive tools using machine and deep learning algorithms} funded by the German Research Foundation (DFG), the \enquote{ML2MT} project from the Volkswagen Stiftung, the \enquote{The Adaptive Mind} project from the Hessian Ministry of Science and Arts (HMWK), the German
Research Center for AI (DFKI) and the Hessian Center for Artificial Intelligence (Hessian.AI). 
It was further supported by the ACATIS Investment KVG mbH project \enquote{Temporal Machine Learning for Long-Term Value Investing}, the Federal Ministry of Education and Research~(BMBF) project KompAKI within the \enquote{The Future of Value Creation -- Research on Production, Services and Work} program (funding number 02L19C150), managed by the Project Management Agency Karlsruhe~(PTKA), and from the EU project EXPLAIN under the BMBF grant 01—S22030D.
It has, furthermore, benefitted from the HMWK project \enquote{The Third Wave of Artificial Intelligence - 3AI}.

\bibliographystyle{ACM-Reference-Format}
\bibliography{bib}

\end{document}

%% file: body.tex
\section{Introduction}
Deep learning (DL) has achieved remarkable advancements in recent years, with state-of-the-art models demonstrating superhuman performance in games like chess \citep{silver2018general} and Go \citep{silver2016mastering} as well as versatile language systems capable of addressing diverse tasks in zero- or few-shot settings \citep{openai2024gpt4technicalreport, touvron2023llama}. Despite these impressive achievements, DL models often rely on \textit{shortcuts}, leading to unexpected failures when applied in real-world settings \citep{GeirhosJMZBBW20}.

Overreliance on specific training artifacts can cause these failures, as they do not generalize to data without these artifacts anymore. This phenomenon can take various forms and occur in many settings:
In medical applications such as diagnosing pneumonia or dementia, models have been shown to depend on irrelevant factors like hospital identifiers or image quality rather than medically significant features \citep{zech2018medical, bottani2023evaluation}. Image classification models have mistakenly relied on embedded photographer tags \citep{lapuschkin2019unmasking} or struggled to identify animals in unusual environments \citep{beery2018recognition}. Models predicting product quality from sensor data in sheet metal manufacturing prioritized irrelevant production speed instead of critical applied forces \citep{kraus2024right}. For sentiment classification, models have used superficial cues such as stop-word distributions instead of focusing on semantically meaningful content \citep{liusie-etal-2022-analyzing}. Even large language models (LLMs) have been found to rely on undesired biases from the input data, negatively impacting their fairness \citep{yuan2024llms, gallegos2024bias}. To discuss the underlying issue in more detail, let us introduce a running example that will serve as a reference throughout this work\footnote{This example follows the waterbird dataset by \citet{sagawa2019distributionally}.}:

\begin{figure*}
    \centering
    \includegraphics[width=\linewidth]{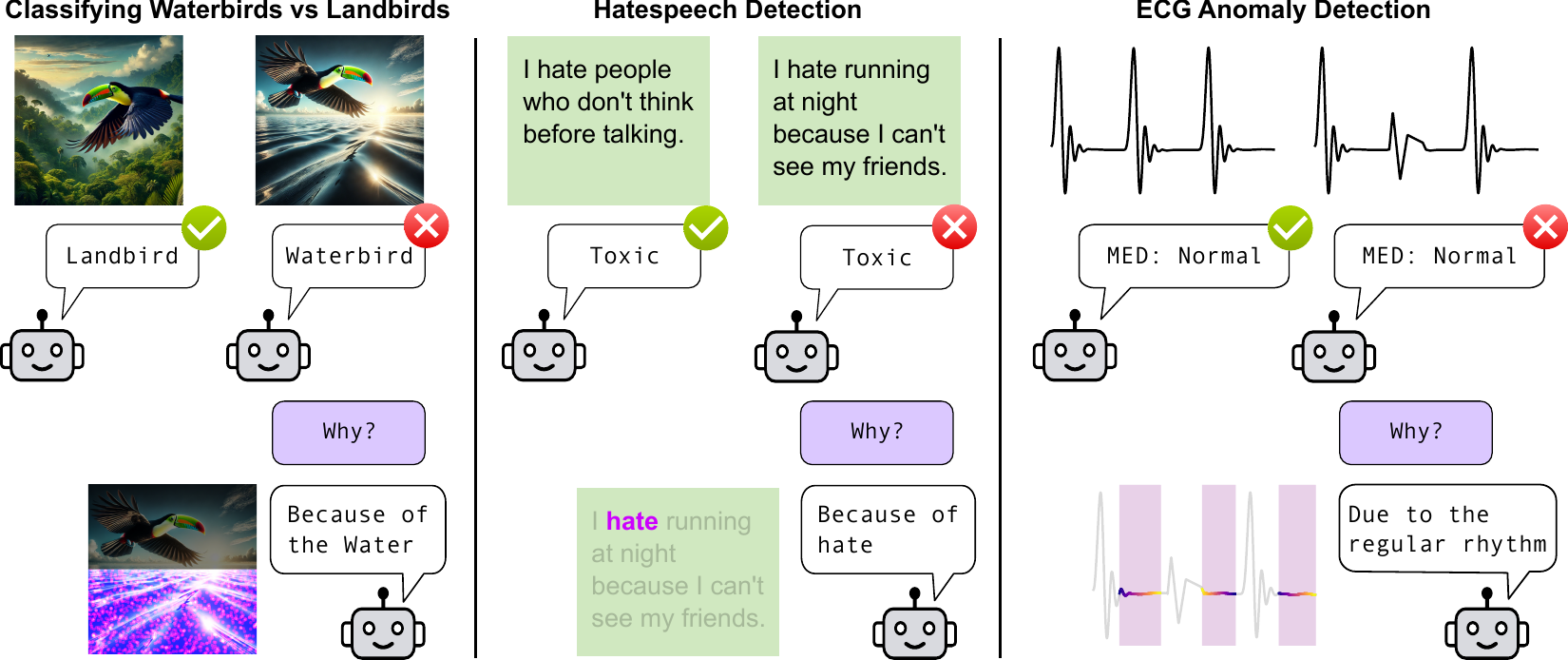}
    \caption{\textbf{Models across different settings are susceptible to shortcuts.}  Models trained on data containing \textcolor{spurious_purple}{\textbf{spurious}} correlations may rely on unintended features for decision-making. These shortcuts can manifest across various domains and tasks, significantly affecting model performance and generalization.}
    \label{fig:intro}
\end{figure*}

\begin{boxD}
For this illustrative example, let us assume that we have a dataset of images depicting various bird species. The goal is to classify these images into two categories: \textbf{landbirds} and \textbf{waterbirds}, based on the birds' characteristics. Naturally, images of landbirds are more likely to feature land backgrounds, while images of waterbirds often include water in the background. However, we want the model to perform its predictions based on the bird itself rather than the environment, as the presence of a landbird in front of a water background does not transform it into a waterbird.
\end{boxD}

When training a model on this waterbird and landbird dataset, it often relies on background features rather than focusing on the birds' relevant characteristics \citep{sagawa2019distributionally}. Similar to the examples discussed earlier, the model follows a shortcut to complete its task. Using the background information can yield high training accuracy, but does not solve the task based on the right reasons. Consequently, the model struggles to generalize to new data where the shortcut is absent, such as landbirds pictured against a water background (cf. \autoref{fig:intro}, left).

This approach to solving tasks is a well-known phenomenon that extends beyond machine learning, appearing, for instance, in animal psychology \citep{samhita2013clever}. In the context of machine learning, it is described using various different terms: \textbf{shortcuts} \citep{geirhos2022imagenet, friedrich2023typology, muller2023shortcut}, \textbf{spurious correlations} \citep{ye2024spurious, sagawa2019distributionally, wu2023discover}, \textbf{Clever Hans behaviour} \citep{lapuschkin2019unmasking, chettri2023clever,anders2022finding}, or \textbf{confounders} \citep{setlurprompting, zhao2020training, zare2022removal}. 
While these terms describe the same fundamental issue, they are often used informally and lack precise definitions, making it challenging to discern their similarities and differences. Moreover, research on the same problem has been independently developed under these different terms, resulting in a fragmented state of the field with many individual research threads. The absence of comprehensive surveys exacerbates this fragmentation, making it challenging for researchers to gain a clear understanding of the field's current state or to find synergies between methods.

In this work, we take a decisive step toward addressing these issues by introducing a clear and formal definition of the underlying fundamental problem centered around the concepts of \textbf{shortcuts} and \textbf{spurious correlations}. We then clarify how this definition relates to other terms like \textbf{Clever Hans behavior}, \textbf{confounders} or \textbf{biases}. Going even one step further, we discuss the connection and overlap between shortcuts and other prominent topics in machine learning, including distribution shifts, causality, or adversarial features. 
In this light, we explore potential sources of shortcuts in machine learning and why models are prone to using them. Given these building blocks, we introduce a taxonomy of shortcut learning, categorizing the rich body of work in this area and bringing research under the various associated terms together. To further facilitate the field's progression, we collect datasets targetting shortcut detection and mitigation and point out open challenges and further research opportunities. 

Overall, the contributions of this work can be summarized as follows:
\begin{itemize}
    \item[(i)] We provide formal definitions of shortcuts, unifying and connecting the terms spurious correlations, Clever Hans behavior, and confounders.
    \item[(ii)] We introduce a taxonomy of shortcut learning - providing an overview of the current state of the field and structuring existing approaches.
    \item[(iii)] Based on this taxonomy, we identify open challenges, like tackling more complex shortcuts, also beyond the typical image classification,  that still need to be addressed.
    \item[(iv)] We provide a comprehensive overview of available datasets that explicitly include shortcuts, facilitating the development of new approaches.
\end{itemize}

The structure of the paper is as follows. In \autoref{sec:definition}, we provide a formal definition of shortcuts and explore their origins. Next, we establish the building blocks of our taxonomy in \autoref{sec:building_blocks} and present the taxonomy itself in \autoref{sec:taxonomy}. In \autoref{sec:detection}, we discuss methods aimed at detecting shortcuts, followed by methods aimed at mitigating these in \autoref{sec:mitigation}. We then collect and review relevant datasets for detection and mitigation in \autoref{sec:datasets}. The paper concludes with a discussion of open challenges in \autoref{sec:outlook} and a summary in \autoref{sec:conclusion}.

\section{Shortcuts and their Origin}
\label{sec:definition}

To establish a foundation for our taxonomy, we first provide a formal definition of shortcuts in the context of machine learning. Following this, we identify under which circumstances they tend to emerge.
We start with establishing a common notation of data, features and correlations as a basis for the definitions.

\paragraph{Data.}
Let us assume there is a ground-truth distribution of observational data $P_\text{gt}(x)$. The samples from the ground-truth $x \sim P_\text{gt}(x)$ consist of multiple features from a joint feature set $F = \{f_i\}_{i=1}^M$, such as raw pixels or higher-level attributes like the wing color of a bird. Unfortunately, this ideal distribution is not available in practice. Instead, one can merely observe a distorted view $P(x) \leftsquigarrow P_\text{gt}(x)$ of the ground truth. A specific dataset $D = \{x_i\}_{i=1}^N$ available for a machine learning task then consists of samples $x_i \sim P(x)$. Importantly, the sampling process follows the distorted distribution $P(x)$ instead of $P_\text{gt}(x)$. In the following, we assume that there are no errors introduced in sampling from $P(x)$ and potential \emph{sampling errors} occur due to the difference between $P(x)$ and $P_\text{gt}(x)$.

\begin{figure}
    \centering
    \includegraphics[width=\linewidth]{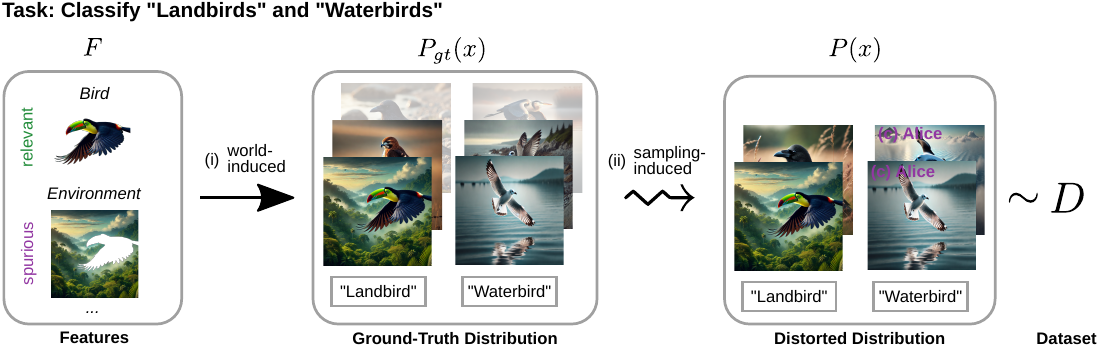}
    \caption{\textbf{Overview where spurious correlations can appear}. Given our established example of classifying birds into landbirds and waterbirds (based on their characteristics), the environment is a spurious feature naturally occurring in the world (i). The distorted ($\rightsquigarrow$) sampling process can then induce spurious correlations through, for example, photographer tags (ii).}
    \label{fig:correlations_overview}
\end{figure}

\paragraph{Features and Correlations.}
Assume we have such a dataset $D$ and a specific task $T: F_\text{input} \xrightarrow{} F_\text{target}$, mapping from an input set $F_{\text{input}} \subseteq F$ to a target set $F_{\text{target}} \subseteq F$ of features, where $F_{\text{input}} \cap F_{\text{target}} = \emptyset$.
We model correlations between features in the dataset through a symmetric correlation function $c: F \times F \xrightarrow{} [0,1]$, indicating that two features $f_i$ and $f_j$ are more correlated if $c(f_i, f_j)$ is closer to $1$. Within the context of machine learning, we are specifically interested in correlations related to the task, i.e., correlations between $f_i \in F_\text{input}$ and $f_j \in F_\text{target}$.
There are different ways to measure whether two features are correlated. The most common one is the Pearson correlation coefficient \citep{sedgwick2012pearson} for linear relationships, but others like Spearman's \citep{myers2004spearman} or Kendall's \citep{abdi2007kendall} rank correlation coefficient can also be used.
In the following, if the specific type of correlation is relevant, it is specifically mentioned.

\paragraph{Spurious Correlations and Shortcuts.}
Given task $T$, some input features are considered relevant to solve the task (in the intended way), which we denote as $F_\text{relevant} \subseteq F_\text{input}$. Correlations $c(f_i, f_j)$ between non-relevant $f_i \notin F_\text{relevant}$ and target $f_j \in F_\text{target}$ are considered \textit{spurious}.
There are two main reasons for spurious correlations to occur.
First (cf. (i) in \autoref{fig:correlations_overview}), the correlation given task $T$ may already be spurious in the ground-truth distribution of observations $P_\text{gt}$. We call such correlations \emph{world-induced}. An example is the environment when classifying waterbirds and landbirds. Waterbirds appear statistically more often on water than landbirds, even in the ground-truth world. As we have established, we want to perform the classification based on the bird's characteristics and not based on the environment. Thus, the correlation between background and landbird/waterbird (originating from $P_\text{gt}(x)$) is spurious.
The second reason (cf. (ii) in \autoref{fig:correlations_overview}) why a correlation $c(f_i, f_j)$ can be spurious originates from the difference between $P_\text{gt}(x)$ and $P(x)$. This so-called \emph{sampling-induced} distortion of the observational data can induce new correlations through, for example, selection bias. For instance, imagine that in $P(x)$, we only have images of waterbirds with a photographer's signature in $D$, whereas no signatures exist in landbird sampels. Then, the correlation between the signature and the waterbird/landbird label is an induced spurious correlation.
A \textit{shortcut} appears when a model uses a spurious correlation as the basis for its decision-making, i.e., relies on spurious instead of relevant features. Both types of spurious correlations, naturally occurring and induced through the distorted sampling process, enable shortcuts.

\subsection{World-Induced Shortcuts}\label{sec:origin_world}
The world is full of correlations, and determining which are relevant and which are spurious can be challenging \citep{spirtes2016causal}. In our example of distinguishing landbirds from waterbirds, we established that the correlation between the bird's habitat (background) and the landbird/waterbird classification is spurious and that a model should focus on the bird's characteristics instead. This approach assumes that we want the model to classify the bird type based on its features. Alternatively, we might want the model to make this decision based on the bird's habitat rather than the bird's features. In that case, the background features are relevant, while the correlation between bird features and the target would be spurious. We call these spurious correlations \textbf{world-induced}.

While, general world knowledge can aid to distinguish between relevant and spurious features, considering the specific task remains important, as it directly influences which features and correlations are relevant \citep{minderer2020automatic}. This origin of shortcuts, i.e., naturally occurring but unwanted correlations, is sometimes also called \textit{world bias}~\citep{suresh2021framework}.

\subsection{Sampling-Induced Shortcuts}
\label{sec:origin_sampling}
When using a dataset for machine learning, it \textit{never} represents the ground-truth world distribution $P_\text{gt}(x)$ precisely. As a result, datasets may contain correlations that do not exist in the ground-truth distribution, which we refer to as \textit{sampling-induced} arising from the distorted sampling process. Generally, these induced correlations are spurious since they do not reflect causal relationships but occur due to errors in the data collection. In the machine learning datasets, such errors can, for example, occur due to careless data scraping \citep{wu2022probing,birhane2024thedark,wei-etal-2024-unveiling}, particularly in large-scale, automatically scraped web datasets. 

At a high level, all sampling-induced spurious correlations can be traced back to \textit{selection bias} \citep{winship1992models}, where the sampling process does not draw samples completely random from the ground-truth distribution. More specifically, errors may arise from over- or underrepresenting specific relationships between features in the data due to sampling bias \citep{mehrabi2021survey} or representation bias \citep{li2019repair}. Moreover, measurement errors can induce other unintended correlations, often referred to as measurement bias \citep{suresh2021framework}.

Most of the time, these induced spurious correlations are not intentionally included in the data, which we refer to as \textit{accidental} spurious correlations. Conversely, it is also possible that shortcuts in datasets are intentional. From an adversarial perspective, dataset manipulations known as data poisoning~\citep{barreno2006poisoning} may introduce spurious correlations that are not present in the ground-truth distribution. In addition to data poisoning, there are benign, intentionally induced shortcuts. Model watermarking~\citep{adi18watermarking,boenisch2021watermarking} is one such application, used to mark model ownership or to link generated content to a specific source~\citep{fernandez23signature,kirchenbauer23llmwatermarking}. Another benign appearance of induced spurious correlations involves research datasets specifically designed to evaluate an algorithm's robustness to such correlations (cf. \autoref{sec:datasets}).

\subsection{Why Do Models Learn Shortcuts?}
In most cases, spurious correlations exist alongside relevant correlations in the data. In these cases, it is theoretically possible for models to rely only on the relevant features instead of the shortcuts. However, models frequently end up using these shortcuts \citep{qiucomplexity}. So, why does this happen?

First of all, a model's task is generally not precisely defined \citep{bachmann2024pitfalls}. For example, whether it's a coarse label image classification or next-token prediction in language models, ML models are typically trained using empirical risk minimization (ERM) on proxy tasks optimized through loss-based optimization. These task formulations do not prevent models from using shortcuts. For instance, if a model is trained to distinguish waterbirds from landbirds, it receives only images and labels without explicit information about what defines each bird type. The broad task definitions do not specify how the task should be solved, thus enabling the model to rely on shortcuts rather than relevant features. It should be noted that while ML relies mostly on these coarse tasks, defining more specific tasks and obtaining the necessary data is a challenge in itself. 

But why do models often seem to favor learning shortcuts over relevant features? One explanation is given by Occam's Razor or simplicity bias \citep{ye2024spurious}. If it is easier to learn the shortcut than the relevant correlations, ERM tends to learn the shortcut \citep{qiucomplexity}. Interestingly, depending on the nature of the dataset and the task, this can even occur when the spurious correlation is not inherently easier to learn \citep{nagarajan2020understanding}. 
Furthermore, models tend to learn shortcuts if the noise in the relevant features is larger than the noise in the spurious features \citep{ye2023freeze, qiucomplexity}. This essentially means that in the given data, the shortcut can be more predictive to solve the task than the relevant features. Overparameterized models are even more susceptible to this tendency, as they may memorize training samples without the shortcut rather than learning relevant features effectively \citep{sagawa2020investigation}.

\section{Establishing the Building Blocks of our Taxonomy}
\label{sec:building_blocks}
\begin{table}[t]
    \centering
    \caption{Overview of existing surveys in the field}
    \small
    \begin{tabular}{lll@{\hspace{5pt}}c@{\hspace{4pt}}c@{\hspace{4pt}}c@{\hspace{4pt}}c}
        \toprule
        Survey & Area & Focus & Definition & Origin & Detection & Mitigation \\
        \midrule
        \citet{friedrich2023typology} & Vision & Shortcut & \xmark & \xmark & \xmark & \cmark \\ 
        Gupta et al. \citep{gupta2024survey} & Vision & Confounder & \xmark & \xmark & \xmark & \cmark \\
        \citet{banerjee2023shortcuts} & Med. Imaging & Shortcut & \xmark & \cmark & \cmark & \cmark\\
        \citet{dogra2024shortcut} & Language & Shortcut & \xmark & \cmark & \cmark & \cmark\\
        \citet{ho2022survey} & Language & Shortcut & \xmark & \xmark & \cmark & \cmark\\
        \citet{GeirhosJMZBBW20} & General & Shortcut & \cmark & \cmark & \xmark & \xmark \\ 
        \citet{ye2024spurious} & General & Spurious Correlation & \cmark & \cmark & \xmark & \cmark \\
        \midrule
        & & Shortcut, Confounder, & & & & \\
        Ours & General & Spurious Correlation, & \cmark & \cmark & \cmark & \cmark \\
        & & Clever Hans & & & \\
        \bottomrule
    \end{tabular}
    \label{tab:surveys}
\end{table}

Although the concept of shortcuts has been known for a long time, it has never been studied from a wholesome perspective. In addition to inconsistent terminology, most work has been very problem-focused or task-specific. However, shortcuts are an important problem that must be tackled from a general perspective. Hence, we establish the first taxonomy of shortcut learning, a detailed overview of the topic to help the research community advance shortcut learning. In the following, we begin with an overview of related work before establishing connections between shortcuts and other machine-learning areas. By integrating these perspectives, we lay the foundation for a unified taxonomy in the next section.

\subsection{Related Work}
While there are numerous works introducing methods to detect or mitigate shortcuts, only a handful of surveys exists \autoref{tab:surveys}. From these, none covers shortcut learning in a comprehensive way. The existing surveys mostly focus on specific areas: vision \citep{friedrich2023typology, gupta2024survey}, medical images \citep{banerjee2023shortcuts} or language \citep{dogra2024shortcut, ho2022survey} and only focus on work under a specific term: shortcut \citep{friedrich2023typology, banerjee2023shortcuts, dogra2024shortcut, ho2022survey, geirhos2022imagenet}, spurious correlation \citep{ye2024spurious} or confounder \citep{gupta2024survey}. While these surveys provide a valuable resource in their specific setting, neither is suitable as a general overview over the field of shortcut learning. To bridge this gap, our work does confine itself to specific target areas and does not focus on specific terms. Instead our introduced taxonomy provides for the first time a general and unifying view on shortcut learning.

Beyond the lack of comprehensive surveys, the terms \textit{shortcut}, \textit{spurious correlation} and \textit{Clever Hans behavior} are generally used informally. To unify the field and understand the differences between individual terms and works, a formal definition is essential. Our definition bases on the description of shortcuts by \citet{GeirhosJMZBBW20}, which describes a shortcut as an unintended solution that still performs well on the training data, so essentially a solution that relies on unintended features. While this captures the essence of shortcuts quite well, it is insufficient to discuss all different terms. Thus, our given definition explicitly covers the origins of shortcuts, relating them with spurious correlations. Further, the more detailed formulization also captures the relation between shortcuts and confounders (cf. \autoref{sec:causality} below).

The most formalized definition of spurious correlations in the field comes from \citet{ye2024spurious}. They define a correlation as spurious if it is between a non-predictive input feature and a target feature. While their approach reflects the perspective of group-robustness optimization \citep{sagawa2019distributionally}, it does not cover all instances of spurious correlations; there can also be spurious correlations with a correlation coefficient of one. The strength of the correlation should not determine whether it is spurious or not. In contrast, our definition explicitly covers the origin of spurious correlations and the resulting shortcuts in the model, thus unifying the concepts of spurious correlations and shortcuts. To further establish the fundaments of our taxonomy, we now connect shortcuts and spurious correlations to the remaining building blocks.

\subsection{From Animal Psychology to Machine Learning: The Clever Hans Phenomenon}
The term \textit{Clever Hans} originates from animal psychology, named after the famous horse Hans that apparently had learned to understand human language \citep{samhita2013clever}. After careful examination, it turned out that Hans learned to rely on the subtle facial expressions of the humans asking the questions and was unable to answer when not seeing the human face. To solve its task, the facial expressions were shortcuts, which Hans learned to utilize.

Based on this story, the term \textit{Clever Hans} has also been adopted outside of animal psychology to describe the usage of unintended cues to solve a task. In the context of machine learning, this corresponds to models learning to rely on shortcuts in the training data instead of the relevant features \citep{lapuschkin2019unmasking}. Clever Hans behavior has been shown to occur in various tasks, e.g., in classification \citep{stammer2021right} or anomaly detection \citep{kauffmann2020clever}. While Clever Hans behavior is seldomly formalized, it corresponds to our definition of a shortcut.

\subsection{Shortcuts and Confounders from a Causality Perspective}\label{sec:causality}

\begin{figure}
    \centering
    \includegraphics[width=0.6\linewidth]{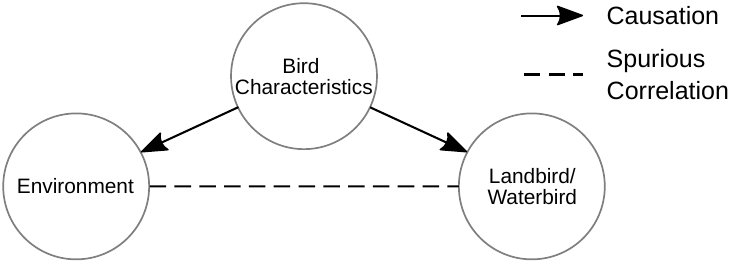}
    \caption{\textbf{Overview of the waterbirds example in the context of causality.} In the data, we have access to the bird's characteristics and its environment and we want to predict whether the bird is a landbird or waterbird. If we assume that the bird's characteristics (i.e., its appearance and abilities) cause both its environment and whether it is a waterbird, environment and the target label are correlated in the data (while not causally related). The common cause (bird characteristics) is called a confounder.}
    \label{fig:causality_example}
\end{figure}

Spurious correlations and shortcuts are inherently connected to the field of causality. One of the most general definitions of a spurious correlation is "a correlation which does not imply causation", which is relevant in causality because it affects learning about genuine causal effects~\citep{pearl2000models}. In the context of causality, the term \textit{confounder} appears regularly as well. To explain how these terms are connected, let us again consider the waterbird example (\autoref{fig:causality_example}). In this context, we have only talked about spurious correlations and intended correlations so far, but we have not talked about actual causal relationships. Let us assume that a bird's characteristics, i.e., its appearance and abilities, cause both its environment (as they influence where a bird usually lives) and whether it is considered a landbird or waterbird. In this case, the bird's characteristics are a confounder: As they cause both the environment and the target label, they are the source of the spurious correlation between both. 

We note that while most machine learning mainly considers identifying correlations and does not model explicit causal relations, this causal perspective can provide valuable insights regarding the origin of spurious correlations. Particularly world-induced spurious correlations can originate from observed or hidden confounders. Moreover, avoiding and addressing confounders is necessary for estimating causal effects, which has a long history in causal inference \citep{miettinen1974confounding} and epidemiology \citep{morabia2011history}. Thus, causal analysis can provide powerful tools to detect and mitigate confounders and the resulting spurious correlations. 

Moreover, the problem of shortcuts can even go beyond confounders in causality. Consider a simple causal graph with three variables: $\mathbf{A} \rightarrow \mathbf{B} \rightarrow \mathbf{C}$. If the relation $\mathbf{B} \rightarrow \mathbf{C}$ is a simple linear correlation, we can predict $\mathbf{B}$ from $\mathbf{C}$. While this can serve as a reasonable correlation basis for predictions in most cases, using $\mathbf{C}$ as a basis to predict $\mathbf{B}$ is not reliable when $\mathbf{B}$ is intervened. On. Similarly, it would be possible to predict $\mathbf{C}$ from $\mathbf{A}$, which again is not reliable if $\mathbf{B}$ is intervened on. So, while addressing confounders is a helpful tool in combatting spurious correlations, it is not necessarily sufficient to mitigate all potential shortcuts.

Finally, it is noteworthy that the term confounders in most ML research is used with a different meaning than the "causal" confounder described above. Some papers rather use "confounder" as a synonym for shortcuts and, in particular, the shortcut features \citep{setlurprompting, zhao2020training, zare2022removal}. To keep terminology clear and avoid confusion, we recommend to rather use the term shortcut in these circumstances and refer to the term confounder for its causal meaning.

\subsection{The Role of Distribution Shift in Shortcut Learning}

When a model has learned a shortcut, it relies on some specific and unintended features to solve its task. This becomes particularly problematic when these features change or are no longer present in the data, as the model then makes incorrect predictions. Distribution shifts can be one of the main reasons why features change, and shortcuts do not work anymore. On the contrary, models that are robust to distribution shifts tend also to be more robust to shortcuts, particularly against sampling-induced ones, as they rely on more robust features \citep{zhou2021examining}. Another way to utilize distribution shifts in the context of shortcut learning is to detect shortcuts. When comparing model performance on the initial data distributions and several shifted versions, potential sources for performance degradation are shortcuts \citep{WilesGSRKDC22}. Overall, the field of shortcut learning can benefit from the research on distribution shifts both to detect potential shortcuts and to develop more robust models, thus mitigating the reliance on shortcuts.

\subsection{Bias as a Potential Origin of Shortcuts}
The term bias is broadly used in discussions about machine learning models and datasets, particularly regarding fairness \citep{mehrabi2021survey}. As bias is used in many different contexts and with different meanings, it is difficult to give an exact definition. Informally, however, bias can be described as a model having a tendency toward specific factors, which are often irrelevant or undesired, such as a credit scoring system relying on race or gender. 
The term also regularly appears in the field of shortcut learning as several works refer to spurious correlations as \textit{dataset biases} (e.g., \citep{kervadec2021roses, seo2022information, yang2024identifying, ragonesi2023learning, luo2022pseudo}).

Further, biases are an important reason why spurious correlations occur in datasets. Inducing spurious correlations through a (distorted) sampling process is often known as sampling or selection bias \citep{winship1992models}, representation bias \citep{li2019repair} or measurement bias \citep{suresh2021framework}. We discuss these biases as an origin for shortcuts in detail in \autoref{sec:origin_sampling}. However, not all biases stem from the data selection and sampling process. Biases that reflect patterns in the world are often termed historical biases \citep{suresh2021framework} and can be seen as correlations in the ground-truth distribution $P_\text{gt}(x)$ that we may not want a model to reflect \citep{liang2021towards, hendricks2018women}. Although the reasons for not reflecting these correlations in a model might be different from the reasons described in \autoref{sec:origin_world}, the differences from a technical perspective are small. This opens the use of shortcut mitigation techniques to address biases, such as mitigating gender bias \citep{zhao2017men}. While this work primarily focuses on techniques to mitigate shortcuts and spurious correlations, many strategies from bias mitigation can be adapted to this purpose. Vice versa, the methods we present to mitigate shortcuts can also be used to reduce biases in machine learning models and datasets. Consequently, there is a large potential for the fields of shortcut and bias mitigation to benefit from each other.

\subsection{Adversarial Features as Shortcuts}\label{sec:adversarial}
Shortcuts are also relevant from an adversarial machine learning perspective, they have a high similarity to backdoor attacks, a common attack on models~\citep{gu2017,chen2017backdoor}.
Backdoor attacks integrate a hidden functionality into by manipulating the training data. When queried with normal inputs, the backdoored model behaves as expected, however adding the trigger to the data activates the model and it produces specific outputs, for example classifying all images as horses. A common strategy to create a backdoor is adding a small number of poisoned samples to the training set, containing a visual trigger like small colored patches. The label of the poisoned samples is set to the desired target class. After training, inputs containing the trigger are always classified as the target class, regardless of their actual content.
In the light of our formalization, we can interpret triggers as adversarial features $f_\text{adv}$ that are added to the set of input features $F_\text{input}$, introducing spurious correlations $c(f_\text{adv}, f_\text{target})$ with target features. These correlations are induced due to a (adversarially) distorted sampling process and not present in $P_\text{gt}(x)$. 
This highlights that the problem of detection and mitigating spurious correlations and the resulting shortcuts is not only relevant for model performance and generalization, but equally as important to maintain security and privacy \citep{yang2022understanding}. On the other hand, there is a lot of work from the security perspective on detecting and mitigating backdoor attacks. These methods can potentially also be used to detect and mitigate shortcuts, and can be a valuable addition to the field.

\section{A Unified Taxonomy of Shortcut Learning}\label{sec:taxonomy}
After establishing the building blocks, we can now introduce our unified taxonomy. Within this taxonomy, we integrate concepts and methods from diverse fields under the overarching term of \textit{shortcut learning}. By unifying research under the terms shortcuts, spurious correlations, Clever Hans behavior, and confounder, we provide a structured and comprehensive perspective on the field. The taxonomy structures approaches first into two main categories, which build upon each other: \textit{shortcut detection} and \textit{shortcut mitigation}.

\begin{figure}[tp]
    \centering
    \includegraphics[width=1\linewidth]{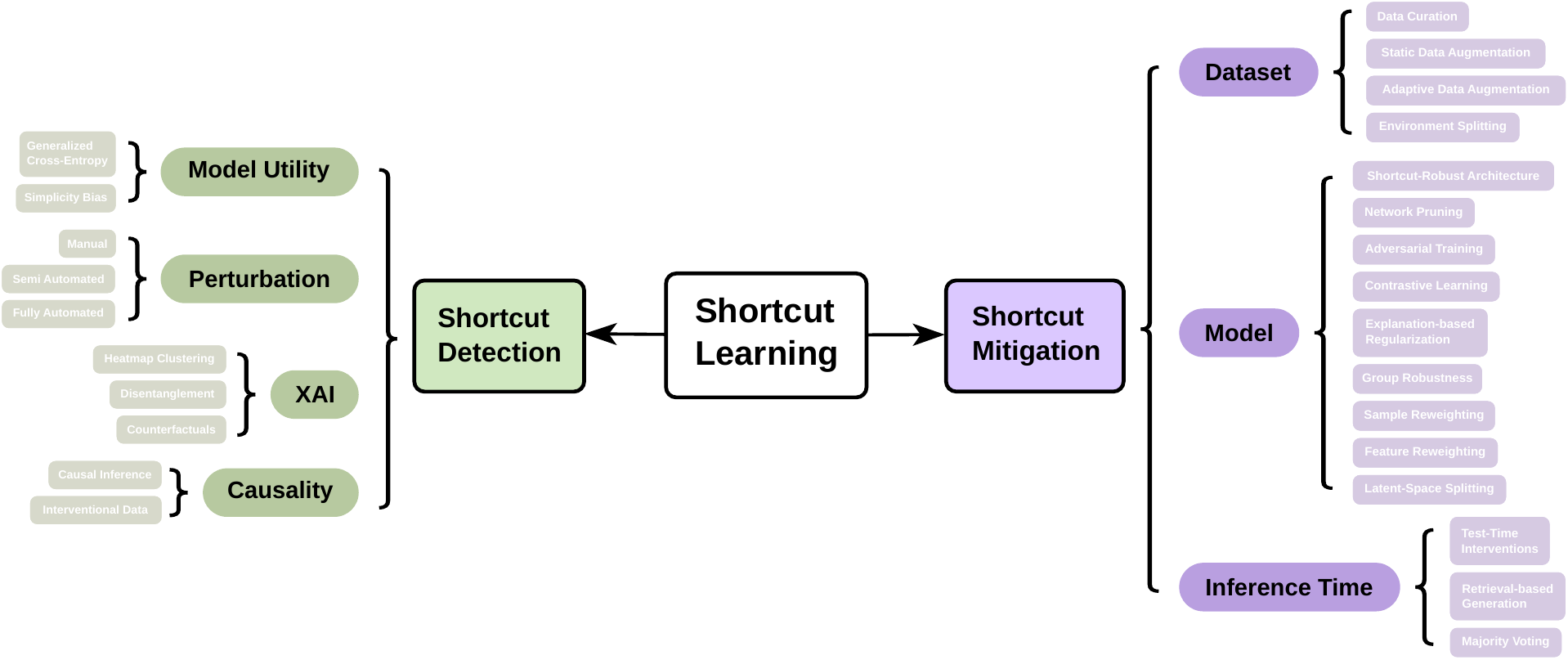}
    \caption{\textbf{Overview of our taxonomy on shortcut learning.} We categorize approaches into the two areas of shortcut detection and shortcut mitigation. Detailed information, including all subcategories, is provided in \autoref{fig:detection} and \autoref{fig:mitigation} in the following sections.}
    \label{fig:taxonomy_overview}
\end{figure}

\paragraph{Shortcut Detection.}
Identifying shortcuts is a critical step in addressing shortcut learning, as mitigating shortcuts first requires an awareness of their presence. Since many mitigation techniques rely on prior knowledge of the shortcuts involved, detection methods are a crucial prerequisite for the mitigation strategies that follow. We categorize shortcut detection methods into four main approaches based on their methodology: detection through evaluating model performance and utility, detection via perturbations, detection using model explanations, and detection through causal analysis. A more detailed breakdown of these categories is illustrated in \autoref{fig:detection}, with an in-depth discussion of each category and its respective methods provided in \autoref{sec:detection}.

\paragraph{Shortcut Mitigation.}
Once shortcuts have been identified, the focus shifts to preventing models from relying on them, which is the goal of shortcut mitigation. This problem can be approached at different stages of the machine learning pipeline: When preparing the datasets, at the model itself, or at inference time. Accordingly, we categorize mitigation methods into these three main groups. A comprehensive breakdown of the subcategories is provided in \autoref{fig:mitigation}, and the specific techniques within each group are discussed in detail in \autoref{sec:mitigation}.

\paragraph{Datasets.} While much of the research in shortcut learning focuses on developing methods for detecting and mitigating shortcuts, datasets play a crucial role in advancing and evaluating these approaches. Although shortcuts exist in many, if not most, datasets, their mere presence is not enough to effectively develop and assess detection and mitigation methods. For meaningful progress, datasets must provide detailed information about the shortcuts they contain, enabling controlled and systematic evaluations. To support this, we present a collection of datasets frequently used in shortcut learning in \autoref{sec:datasets}. The datasets are not explicitly listed in the taxonomy overview \autoref{fig:taxonomy_overview}, as they are implicitly part of shortcut detection and mitigation.

This taxonomy offers a structured overview of the field, serving as a valuable resource for both practitioners and researchers. For practitioners, it compiles methods for shortcut detection and mitigation, facilitating their application in practical settings. For researchers, it provides a comprehensive summary of the current state of the field, shedding light on critical challenges that require further investigation. The following sections explain the two parts of shortcut detection and mitigation in detail.

\section{Detection}\label{sec:detection}

\begin{wrapfigure}{r}{0.6\linewidth}
    \centering
    \vspace{-1.15cm}
    \includegraphics[width=0.75\linewidth]{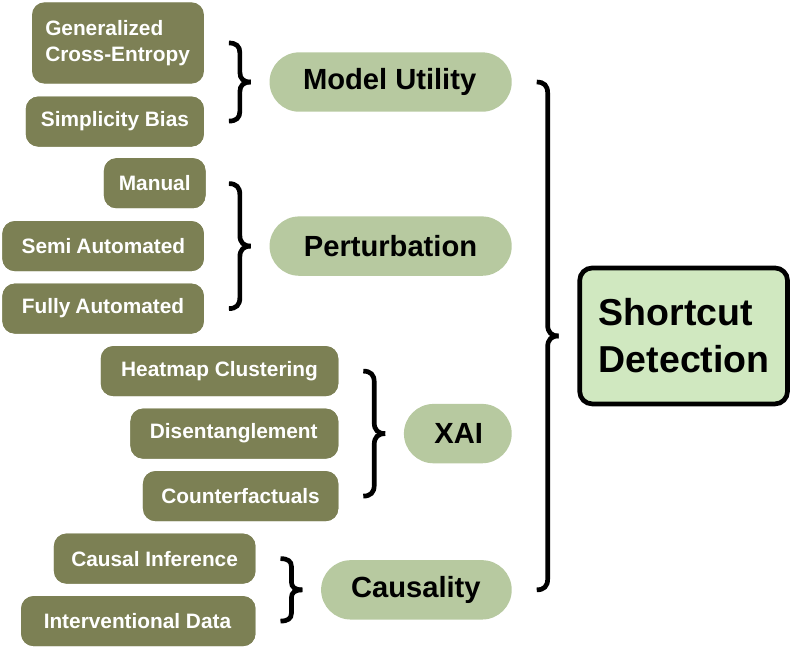}
    \caption{\textbf{Taxonomy of shortcut detection approaches.}
        A comprehensive breakdown of shortcut detection methods, organized into methodological subcategories.}
    \vspace{-0.5cm}
    \label{fig:detection}
\end{wrapfigure}

To address shortcuts in the data, it is first necessary to recognize their presence. As discussed in the previous sections, it is often challenging to decide what features are spurious, as this decision depends on the task and its intended solution. To circumvent this, some methods aim to provide effective tools for domain experts to understand what features a model is relying on and let them 
decide whether these features are spurious or not. To design automatic detection methods without explicit human interactions, other methods pose some more specific assumptions about the nature of shortcuts in the data, such as the presence of minority groups (i.e., a small number of samples where spurious features are absent). The assumptions of the different methods impact the way how shortcuts can be detected and addressed. 
Based on existing literature, we have identified four main categories of detection methods: assessing model utility, detecting shortcuts via perturbations, detection using XAI techniques, as well as causality-based methods (cf. \autoref{fig:detection}).

\subsection{Detection via Model Utility}
\label{sec:det:utility}

The underlying assumption of the following works is that shortcuts are easier to learn than the relevant features. Additionally, they often assume that there are some samples without the shortcut present. The methods then use these assumptions to detect samples with and without spurious features. 

\subsubsection{Generalized Cross Entropy.}
Generalized cross-entropy (GCE) is a specific way to promote the learning of easy-to-learn features. \citet{luo2022pseudo} and \citet{nam2020learning} train an additional detector with a GCE loss to distinguish between easy-to-learn and hard-to-learn samples. With the assumption that shortcuts are easier to learn, this detector can provide pseudo-labels to use for shortcut mitigation. 

\subsubsection{Simplicity Bias.}
\citet{yang2024identifying} propose SPARE, which aims to identify shortcuts early in training by leveraging simplicity bias. Their method clusters network outputs from early epochs to separate majority and minority groups, assuming that shortcuts are already learned early during training. This clustering is followed by importance sampling to mitigate these correlations.
\citet{labonte2024towards} suggest training multiple models with strong regularization techniques such as dropout and early stopping, identifying non-shortcut samples by finding those consistently correctly classified across models. 
\citet{yenamandra2023facts} first amplify shortcuts by training with a large weight decay rate, followed by correlation-aware clustering to discover shortcut-conflicting slices of data. This approach is suitable for scenarios where, at most, one spurious attribute exists. Similarly, \citet{dagaev2023too} use a low-capacity network to detect easy-to-learn features that are likely to be spurious. These can then be down-weighted during training of a high-capacity network to encourage the focus on relevant features.

\subsubsection{Miscellaneous.}
\citet{adnan2024detecting} detect shortcuts using mutual information between input and learned representations. This involves training an infinite-width model using neural tangent kernels \citep{jacot2018neural} to compute mutual information for both training and out-of-distribution (OOD) test data. If mutual information for the test data is significantly lower than for the training data, it suggests the presence of shortcuts, highlighting correlations that may not generalize well beyond the training distribution.
While initially designed to detect backdoor triggers, SCALE-UP \citep{GuoLCG0023} could also be used to detect shortcuts. This method analyzes the consistency of predictions when scaling the input data to detect potential irregularities.

\begin{table}[tp]
    \centering
    \small
    \caption{Overview of different shortcut detection methods. Methods are sorted based on category and subcategory. More information about the methods is provided in Sec.~\ref{sec:det:utility}--\ref{sec:det:causal}.}
    \resizebox{\textwidth}{!}{
    \begin{tabular}{l@{\hspace{1pt}}llll}
        \toprule
        & & Subcategory & Method & Description \\ 
        \midrule 
        \multirow{7}{*}{\rotatebox{90}{\textit{Model Utility}}} & \multirow{7}{*}{\rotatebox{90}{\textit{(Sec.~\ref{sec:det:utility})}}} & Simplicity Bias & \citet{yang2024identifying} & Identify shortcuts early during training \\
        & & Simplicity Bias & \citet{labonte2024towards} & Multiple regularized models as detector ensemble \\
        & & Simplicity Bias & \citet{yenamandra2023facts} & Amplify shortcut detection through weight-decay \\
        & & Simplicity Bias & \citet{dagaev2023too} & Low-capacity network as shortcut detector \\
        & & Generalized CE & \citet{luo2022pseudo} & Shortcut detector with GCE loss\\
        & & Generalized CE & \citet{nam2020learning} & Shortcut detector with GCE loss\\
        & & Miscellaneous & \citet{adnan2024detecting} & Mutual feature information as shortcut indicator \\
        & & Miscellaneous & \citet{GuoLCG0023} & Prediction consistency when scaling input features \\ 
        \midrule
        \multirow{5}{*}{\rotatebox{90}{\textit{Perturbation}}} & \multirow{5}{*}{\rotatebox{90}{\textit{(Sec.~\ref{sec:det:pertubation})}}} & Manual & \citet{chettri2023clever} & Perturbations in voice spoofing detection \\
        & & Manual & \citet{sturm2014simple} & Method of irrelevant transformations \\
        & & Semi Automated & \citet{agarwal2020towards} & Steering semantic automated image manipulations \\
        & & Semi Automated & \citet{brown2023detecting} & Perturb encodings of sensitive attributes \\
        & & Fully Automated & \citet{wang2023neural} & Frequency-based perturbations \\
        \midrule
        \multirow{10}{*}{\rotatebox{90}{\textit{XAI}}} & \multirow{10}{*}{\rotatebox{90}{\textit{(Sec.~\ref{sec:det:xai})}}} & Heatmap Clustering & \citet{lapuschkin2019unmasking} & Spectral clustering of LRP explanations \\
        & & Heatmap Clustering & \citet{SchramowskiSTBH20} & Spectral clustering of explanations\\
        & & Heatmap Clustering & \citet{moayeri2023spuriosity} & Heatmaps of adversarially trained networks \\
        & & Disentanglement & \citet{chormai2024disentangled} & Separate explanations into conceptual concepts \\
        & & Disentanglement & \citet{carter2021overinterpretation} & Sufficient Input Subsets for predictions \\
        & & Disentanglement & \citet{muller2024shortcut} & Disentangle meaningful features via VAE \\
        & & Disentanglement & \citet{bykov2023finding} & Compare functional \& concept-based distances \\
        & & Disentanglement & \citet{szyc2021checking}  & Salient features within bounding boxes \\
        & & Counterfactuals & \citet{DegraveJL21} & Counterfactuals for COVID-19 lung images \\
        & & Counterfactuals & \citet{sikka2023detecting} & Counterfactual-based neuron activations for trojan triggers \\ 
        \midrule
        \multirow{3}{*}{\rotatebox{90}{\textit{Causal}}} & \multirow{3}{*}{\rotatebox{90}{\textit{(Sec.~\ref{sec:det:causal})}}} & Interventional data & \citet{kumar2024causal} & Estimate causal relationships with interventional data \\
        & & Causal Inference & \citet{zheng2022causally} & Model data generation to detect shortcuts \\
        & & Causal Inference & \citet{karlsson2024detecting} & Identify hidden confounders between multiple environments \\ 
    \bottomrule
    \end{tabular}}
    \label{tab:detection_overview}
\end{table}

\subsection{Perturbation-Based Detection.}
\label{sec:det:pertubation}
Perturbation-based detection methods examine how model performance changes when data is systematically modified, revealing any reliance on potential spurious features. In general, we can differentiate between automatic augmentations, semi-automatic generations, and manual approaches.

\subsubsection{Manual Perturbations.}
Several works utilize domain knowledge to craft these perturbations manually. For example, \citet{chettri2023clever} focus on detecting shortcuts in voice spoofing detection by evaluating a model's performance on both original and augmented versions of the dataset. The augmented data includes adding or removing artifacts, such as specific audio features that do not correlate with genuine or spoofed labels but may serve as shortcuts. The difference in model performance serves as a coarse detection mechanism for shortcuts.
A similar methodology is employed by \citet{sturm2014simple} in the context of music information retrieval systems. In their work, they propose the \enquote{method of irrelevant transformations}, modifying data using changes that should not affect the target variable (e.g., applying slight equalization or cropping irrelevant parts of audio recordings). Changes in model performance reveal dependencies on dataset-specific shortcuts that might be unrelated to actual musical content.

\subsubsection{Semi Automated Perturbations.}
In contrast, \citet{agarwal2020towards} leverage automated semantic image manipulations to assess model robustness in Visual Question Answering~(VQA). By steering the generation process of a generative adversarial network (GAN), they modify certain image features, allowing them to test for model consistency across different versions. Unlike manually crafted augmentations, their use of GANs provides an automated yet contextually informed approach to detecting shortcuts.
\citet{brown2023detecting} propose ShorT, which specifically intervenes on model encoding of sensitive attributes (e.g., age or race). By systematically altering these encodings and assessing their impact on model performance and fairness, ShorT identifies whether they act as shortcuts, ensuring fairer AI predictions in medical applications.

\subsubsection{Fully Automated Perturbations.}
\citet{wang2023neural} provide a fully automatic process to reveal frequency-based shortcuts in networks for image classification. They perturb the data by sequentially removing specific frequency bands to assess their importance. This enables ranking frequencies by their effect on model loss and helps identify whether models over-rely on particular frequency components. They further test model performance with only the top 5\% of frequencies, detecting whether neural networks only focus on narrow aspects of the data spectrum, which they deem a likely shortcut.

There also exist other methods related to the detection of shortcuts in the context of adversarial feature detection \citep{XueWWZWL23, GaoXW0RN19} using intentional adversarial perturbations. By adding a universal adversarial perturbation to an image and comparing the model's predictions on the perturbed and unperturbed images, these methods can identify backdoors. While some of them need to know the trigger size \citep{XuWLBGL21, QiaoYL19}, others need many clean images \citep{LiuXS17} or multiple trained models \citep{ZhangGMRS21}. 

Overall, perturbation-based methods for shortcut detection provide a direct way to explore model behavior under systematically altered data conditions. However, these methods generally require domain expertise to identify relevant perturbations or artifacts, making them highly task-dependent.

\subsection{Detection via XAI}
\label{sec:det:xai}
Deciding which features are spurious and which are relevant is difficult without domain knowledge, so a suite of methods has been designed to help domain experts identify shortcuts. To analyze the focus of the model, they use common XAI techniques \citep{jin2023shortcutlens}. However, just using existing libraries (e.g., AIX360~\citep{arya2022ai}, XAITK~\citep{hu2023xaitk}, InterpretML~\citep{nori2019interpretml} or
Captum~\citep{kokhlikyan2020captum}) to detect shortcuts out of the box remains challenging, particularly for more complex shortcuts. Thus, this section covers methods that facilitate the analysis of model explanations to detect shortcuts. 

\subsubsection{Heatmap Clustering.}
\citet{lapuschkin2019unmasking, SchramowskiSTBH20} both cluster explanation heatmaps and then present them to a human for evaluation, which is extended to enable automation using context activation vectors (CAV)~\citep{anders2022finding}.
Similarly, \citet{moayeri2023spuriosity} present heatmaps of multiple adversarially trained networks or linear layers (on top of a pretrained encoder) to humans, who then decide whether the identified features are reasonable or not.

\subsubsection{Disentanglement.}
Several works try to disentangle either the explanation or input space to find shortcuts on a discretized level.
\citet{chormai2024disentangled} aims to separate explanations into components that represent different concepts, while \citet{carter2021overinterpretation} directly compute input subsets that are sufficient for a high-confidence prediction and \citet{muller2024shortcut} utilize a VAE to disentangle meaningful features in a given dataset. These methods enable a domain expert to detect potential shortcuts on a higher conceptual level. 
Furthermore, \citet{bykov2023finding} identify shortcuts between output representations by comparing functional and concept-based distances using extreme activations and Wu-Palmer metrics from WordNet. \citet{szyc2021checking} measure how much of the model's salient features (from the saliency map) fall within a bounding box.

\subsubsection{Counterfactual Generation.}
\citet{DegraveJL21} use saliency maps to let radiologists decide if the model focuses on spurious features in lung images, such as laterality markers, arrows, annotations unique to the dataset, and image borders. Then, they generate counterfactuals of different COVID-19 statuses to highlight both relevant and spurious features.
By utilizing neuron attributions to identify ghost neurons that exhibit distinct behavior when Trojan triggers are present, \citet{sikka2023detecting} demonstrate how counterfactual-based neuron excitation can reveal accuracy drops, enabling detection of malicious behaviors.

\subsection{Causality-Based Detection}
\label{sec:det:causal}
Causality-based detection methods aim to estimate causal effects to detect the impact of spurious correlations in machine learning models. These approaches typically leverage knowledge of the data generation process through causal inference techniques to identify the resulting shortcuts.

\subsubsection{Interventional Data.}
For example, \citet{kumar2024causal} introduce causal effect regularization, which aligns model predictions with estimated causal effects by using interventional distributions to regularize the model. This method assumes the availability of interventional data to estimate causal relationships accurately. 

\subsubsection{Causal Inference.}
\citet{zheng2022causally} propose a two-step approach that leverages auxiliary labels to identify and remove shortcuts. Their method relies on a causal directed acyclic graph to represent the data generation process, allowing the model to detect shortcuts.
\citet{karlsson2024detecting} focus on identifying hidden confounders using data from multiple environments with different levels of confounding. By analyzing conditional independencies across these environments, they detect hidden shortcuts that affect the relationships. This approach assumes a well-understood causal mechanism underlying the data.

Overall, while causality-based methods offer a systematic approach to enhance model robustness by addressing spurious correlations, they often depend on strong assumptions, such as access to interventional data or detailed causal structures, which can limit their practical applicability.

\subsection{Open Challenges in Shortcut Detection}
\label{sec:det:future}
The previous overview highlights that many shortcut detection methods still rely, at least partially, on human input. For example, most XAI-based methods aim to refine model explanations, making it easier for humans to identify shortcuts. Similarly, perturbation-based approaches often depend on human effort to select irrelevant features for perturbation.

As discussed in \autoref{sec:definition}, determining whether a feature is relevant or spurious is inherently complex and particularly challenging without access to commonsense knowledge. Given these challenges, it is natural to involve humans in the final decision regarding the presence of shortcuts. However, as the field progresses, the practicality and scalability of these methods must be considered. While human interaction can offer valuable insights, it is crucial to limit the extent of this involvement, as users are often reluctant to label or validate large numbers of samples \citep{amershi2014power}. Therefore, a key direction for future research is to maximize the efficiency of human interactions in shortcut detection methods.

Additionally, many detection methods are built on specific assumptions. For instance, utility-based approaches often assume that shortcut features are easier to learn than relevant features. XAI-based methods presume that shortcuts can be revealed through explanation techniques, while causality-based approaches may assume the availability of counterfactual or interventional data. However, these assumptions are rarely made explicit, making it difficult to assess the true capabilities and limitations of a given method. To address this, future work should state the underlying assumptions more clearly. Moreover, evaluating existing methods in more diverse and comprehensive scenarios to test these assumptions is essential for assessing the current state of the field.

\section{Mitigation}
\label{sec:mitigation}
\begin{wrapfigure}{r}{0.5\linewidth}
    \centering
    \vspace{-0.45cm}
    \includegraphics[width=\linewidth]{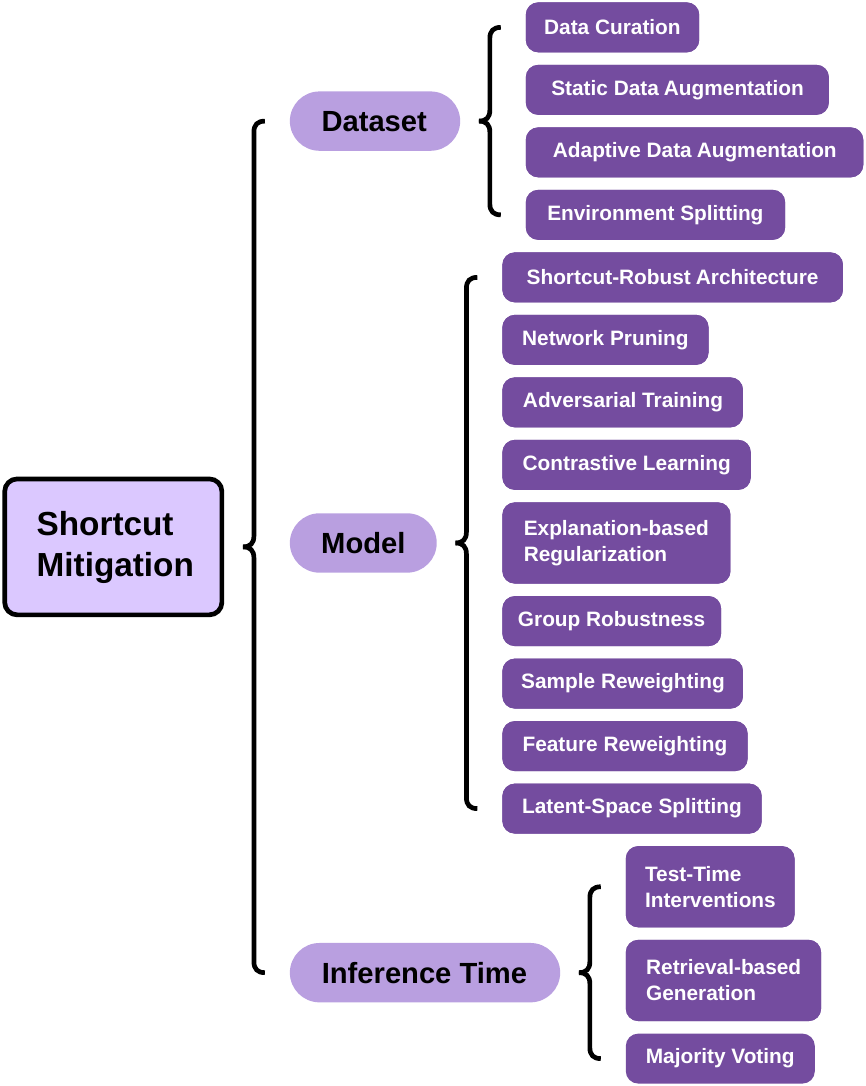}
    \caption{\textbf{Detailed taxonomy of shortcut mitigation approaches.} 
    An in-depth representation of shortcut mitigation strategies, 
    categorizing methods along their main application level and their mitigation strategy.}
    \label{fig:mitigation}
    \vspace{-0.7cm}
\end{wrapfigure}

The methods discussed in the previous section help to identify whether shortcuts are present in the dataset, but detection alone is not sufficient to mitigate shortcut effects on the model. To address this, a wide range of methods is available. We categorize the methods based on their main level of application (cf. \autoref{fig:mitigation}). They can either apply at the dataset level (\autoref{sec:mitigation_data}), at the model level (\autoref{sec:mitigation_model}) or at inference time (\autoref{sec:mitigation_inference}). In the following, we cover each area in more detail.

\begin{table}[tp]
    \centering
    \scriptsize
    \caption{Overview of different shortcut mitigation methods, sorted by category and subcategory.}
    \resizebox{\textwidth}{!}{
    \begin{tabular}{l@{\hspace{1pt}}llll}
        \toprule
        & & Subcategory & Method & Description \\ 
        \midrule 
        \multirow{18}{*}{\rotatebox{90}{\textit{Dataset}}} & \multirow{18}{*}{\rotatebox{90}{\textit{(Sec.~\ref{sec:mitigation_data})}}} & Data Curation & \citet{ahmed2021covid19} & Cleaning shortcuts in Covid-19 Data\\
        & & Data Curation & \citet{portal2022new} & Patch removal in chest X-rays \\
        & & Static Data Aug. & \citet{teso2019explanatory} & Mask shortcuts with noise\\
        & & Static Data Aug. & \citet{plumb2021finding} & Mask shortcut features\\
        & & Static Data Aug. & \citet{nauta2021uncovering} & Inpaint shortcuts in medical images with GANs\\
        & & Static Data Aug. & \citet{kwon2024learning} & Augment foregrounds with random backgrounds\\
        & & Static Data Aug. & \citet{noohdani2024decompose} & Augment CAM foregrounds with random backgrounds\\
        & & Static Data Aug. & \citet{mao2021generative} & Augmented samples with style transfer\\
        & & Static Data Aug. & \citet{wu2023discover} & MixUp-based data augmentation\\
        & & Static Data Aug. & \citet{wang2023dfm} & Frequency-based data augmentation\\
        & & Static Data Aug. & \citet{wang2021robustness} & Augment using language antonyms\\
        & & Static Data Aug. & \citet{gowda2021pulling} & Causal bootstrapping\\
        & & Adaptive Data Aug. & \citet{seo2022information} & Stochastic label noise\\
        & & Adaptive Data Aug. & \citet{lee2021learning} & Random shortcut feature swapping\\
        & & Adaptive Data Aug. & \citet{anders2022finding} & Latent shortcut feature augmentation\\
        & & Environment Splitting & \citet{zare2022removal} & Split dataset via shortcut presence\\
        & & Environment Splitting & \citet{creager2021environment} & Split dataset with worst-case environments\\
        \midrule
        \multirow{48}{*}{\rotatebox{90}{\textit{Model}}} & \multirow{48}{*}{\rotatebox{90}{\textit{(Sec.~\ref{sec:mitigation_model})}}} & Architectures & \citet{li2022discover} & Alternating classifier and shortcut discoverer\\
        & & Architectures & \citet{liu2022contextual} & Subtract learned shortcut from original features\\
        & & Architectures & \citet{wang2022learning} & Explicitly fit unknown concepts\\
        & & Architectures & \citet{luo2021rectifying} & Combine shortcut clustering with few-shot classification\\
        & & Architectures & \citet{ma2023eye} & Adapt attention of vision tranformers\\ 
        & & Architectures & \citet{li2023whac} & Replace the last layer with an ensemble\\
        & & Network Pruning & \citet{linhardt2024preemptively} & Preemptive pruning based on explanations\\
        & & Adversarial Training & \citet{zhao2020training} & Adversarial training to ignore shortcuts\\
        & & Adversarial Training & \citet{adeli2021representation} & Minimize statistical dependence between shortcuts and normal features\\
        & & Adversarial Training & \citet{robinson2021deep} & Adversarial training for medical data\\
        & & Adversarial Training & \citet{ren2022dice} & Adversarial training with replay buffer\\
        & & Adversarial Training & \citet{minderer2020automatic} & Adversarially trained lens to remove shortcuts\\
        & & Contrastive Learning & \citet{teney2020learning} & Use contrastive learning to mitigate shortcuts\\
        & & Contrastive Learning & \citet{wang2022counterexample} & Emphasize the role of counterexamples\\
        & & Contrastive Learning & \citet{yang2023mitigating} & Combine vision and language\\
        & & Explanation Regularization & \citet{ross2017right} & Penalize explanations on shortcut features\\
        & & Explanation Regularization & \citet{SchramowskiSTBH20} & Penalize Grad-CAM explanations\\
        & & Explanation Regularization & \citet{mustafa2024unmasking} & Jacobian saliency map regularization\\
        & & Explanation Regularization & \citet{stammer2021right} & Regularize conceptual explanations\\
        & & Explanation Regularization & \citet{dreyer2024hope} & Regularization with concept activation vectors\\
        & & Explanation Regularization & \citet{bassi2024improving} & Penalize background explanations\\
        & & Group Robustness & \citet{sagawa2019distributionally} & Group DRO\\
        & & Group Robustness & \citet{zhou2021examining} & DRO with a set of joint groups and features\\
        & & Group Robustness & \citet{nam2022spread} & Use pseudo group labels for DRO\\
        & & Group Robustness & \citet{chakraborty2024exmap} & Cluster samples to obtain group labels \\
        & & Sample Reweighting & \citet{luo2022pseudo} & Weighted Softmax Function\\
        & & Sample Reweighting & \citet{kirichenko2022last} & Finetune a model on weight-balanced data\\
        & & Sample Reweighting & \citet{izmailov2022feature} & Finetune a model on weight-balanced data\\
        & & Sample Reweighting & \citet{nam2020learning} & Shortcut Detector module as basis for Reweighting\\
        & & Feature Reweighting & \citet{chen2022does} & Downweight features based on group information\\
        & & Feature Reweighting & \citet{zhang2023robustness} & Reweight input \& high-level features \\
        & & Feature Reweighting & \citet{yang2024identifying} & Reweight early and continuously during training\\
        & & Feature Reweighting & \citet{asgari2022masktune} & Finetune with important features completely masked\\
        & & Feature Reweighting & \citet{holstege2023removing} & Find latent spurious subspaces and reweight\\
        & & Latent-Space Splitting & \citet{yang2022chroma} & Mutual information criterion for splitting\\
        & & Latent-Space Splitting & \citet{fay2023avoiding} & Mutual information criterion for splitting\\
        & & Latent-Space Splitting & \citet{wang2024navigate} & Start training with prefect shortcut-encoding latent space\\
        & & Other Loss-Based & \citet{venkataramani2024causal} & Finetune feature extractor with alignment loss\\
        & & Other Loss-Based & \citet{kumar2024causal} & Regularize causal effect of features on output\\
        & & Other Loss-Based & \citet{tiwari2022robust} & Regularize causal strength between features\\
        & & Other Loss-Based & \citet{veitch2021counterfactual} & Causal approximation of counterfactual inference\\
        & & Miscellaneous & \citet{ragonesi2023learning} & Address shortcuts through meta-learning \\
        & & Miscellaneous & \citet{wang2021counterfactual} & Causality-based training with a structural causal model\\
        & & Miscellaneous & \citet{makar2022causally} & Combine sample weighting and loss regularization\\
        & & Miscellaneous & \citet{arefin2024unsupervised} & Learn concepts unsupervised to solve tasks without shortcuts\\
        & & Miscellaneous & \citet{deng2024robust} & Progressively expand dataset during training\\
        & & Miscellaneous & \citet{moayeri2023spuriosity} & Finetune models on samples with low shortcut occurrence\\
        & & Miscellaneous & \citet{labonte2024towards} & Disagreement between multiple strongly regularized models\\
        
        \midrule
        \multirow{4}{*}{\rotatebox{90}{\textit{Inference}}} & \multirow{4}{*}{\rotatebox{90}{\textit{(Sec.~\ref{sec:mitigation_inference})}}} & Test-Time Interventions & \citet{steinmann2023learning} & Correct shortcut concepts via interventions \\
        & & Test-Time Intervention & \citet{sun2024exploring} & Adapt LLM prompts during inference \\
        & & Retrieval-based Generation & \citet{friedrich2023revision} & Modifying LLM text generation via preference retrieval\\
        & & Majority Voting & \citet{sarkar2020suppression} & Majority voting over noisy input samples\\
    \bottomrule
    \end{tabular}}
    \label{tab:mitigation_overview}
\end{table}

\subsection{Mitigation at Dataset Level}
\label{sec:mitigation_data}

The first set of strategies focuses on mitigating shortcuts directly at the dataset level. They aim to remove the underlying spurious correlations, preventing a model from learning them by ensuring they are no longer present. These methods address both sampling-induced and naturally occurring spurious correlations within the dataset. This section covers all methods that directly modify the available data, while approaches like sample reweighting are discussed in later sections.

\subsubsection{Data Curation.} In real-world machine learning applications like healthcare, datasets typically go through thorough preprocessing steps, including data cleaning and feature selection \citep{hassler2019importance}. By addressing shortcuts during this process, many spurious correlations, whether naturally occurring or introduced through sampling, can be mitigated \citep{ahmed2022achieving, portal2022new}. However, this process is inherently domain and task-specific, so there is no general method for data curation against shortcuts.

\subsubsection{Static Data Augmentation.} Beyond careful selection and filtering of the available data, static data augmentation covers approaches that augment the existing dataset prior to training with new samples to break existing spurious correlations \citep{shen2022augmentation}. When segmentations of the spurious features for the input images are available, \citet{teso2019explanatory} and \citet{plumb2021finding} remove the shortcut by masking these features from the image. One step further, \citet{nauta2021uncovering} train GANs to inpaint medical images and remove existing shortcuts. Similar approaches have also been used to remove backdoor triggers by generating synthetic variations of images with diffusion models \citep{struppek23leveraging}.

Instead of removing the spurious features altogether, a different approach is to create new samples to balance the relationship between target and spurious features, thus removing the spurious correlation. Following this idea, \citet{kwon2024learning} combine segmented foreground objects and arbitrary image backgrounds to mitigate shortcuts tied to the image background. \citet{noohdani2024decompose} avoid the need for annotations and instead use class-activation mappings (CAM) to obtain foreground and background segmentations. To obtain new samples that are more natural than arbitrary combinations of foregrounds and backgrounds, \citet{mao2021generative} first generate augmented versions of the original data with GANs, which are then transferred to the original dataset using neural style transfer. \citet{wu2023discover} perform mixup-based data augmentation utilizing high-level conceptual information about shortcuts to create an augmented dataset with balanced occurrences. 

Shortcuts can also be hidden in different input representations, for example, in the frequency domain. \citet{wang2023dfm} show that models utilize frequency shortcuts and utilize data augmentation to mitigate them. They create inputs where important frequencies of random classes are removed to increase the model's robustness to frequency shortcuts and its resilience against adversarial attacks.
While many data augmentation techniques are designed for images, there is also some work for textual data. \citet{wang2021robustness} generate augmented data with antonyms of relevant features as counterfactuals to extend existing datasets. 
Independent of the data domain, \citet{gowda2021pulling} propose causal bootstrapping, a data augmentation technique that can be used for partially observed or hidden sources of shortcuts. It aims to find the interventional distribution via do-calculus and then derive suitable weights for bootstrapping.

\subsubsection{Adaptive Data Augmentation.} Unlike static approaches from the previous section, adaptive data augmentation techniques augment the dataset not before but throughout the training process. \citet{seo2022information} demonstrate that introducing stochastic label noise, i.e., changing sample labels with certain probabilities in each mini-batch, can weaken spurious correlations. 
\citet{lee2021learning} first use a shortcut prediction network to find important features (which are assumed to be spurious). Then, the values of these features are randomly swapped with other data points during model training. \citet{anders2022finding} localize known shortcuts in linear separable directions of intermediate layers of a model. This latent representation is then used during training to augment the data and balance the influence of shortcuts.

\subsubsection{Environment Splitting.} Invariant risk minimization (IRM) \citep{arjovsky2019invariant} is a training strategy for data stemming from different environments (for example, chest radiographs from different hospitals). In the context of shortcuts, IRM is sometimes used when the different environments can be the source of spurious correlations in the data. To apply IRM, it is necessary to split the dataset according to the different environments. \citet{zare2022removal} split the dataset according to the influence of present shortcuts, while \citet{creager2021environment} try to find the worst-case environments for existing classifiers.

\subsection{Mitigation at Model Level}
\label{sec:mitigation_model}

This section covers methods that tackle shortcuts at the model level. The main idea in this category is to accept the presence of spurious correlations in the dataset but to prevent the model from learning them. Many of the methods that adapt the model training process can be either applied during the full training or in a separate finetuning phase. 

\subsubsection{Shortcut-Robust Architectures.}
Several works introduce specific architectures more robust to shorcuts. \citet{li2022discover} combine a classifier and a discoverer module that are trained alternatingly. The discoverer identifies shortcuts within the classifier, allowing the classifier to be adjusted and regularized to mitigate these shortcuts. \citet{liu2022contextual} use a classifier to explicitly learn shortcuts in the input, which are then subtracted from the original features (obtained via a backbone model) to obtain a representation without shortcuts. \citet{wang2022learning} train a concept-bottleneck model that explicitly encodes known concepts and a residual model to fit potential unknown concepts. The latter is regularized to avoid correlations between unknown and known concepts, allowing the encoded concepts to help avoid shortcuts. \citet{luo2021rectifying} introduce COSOC, an architecture tailored for few-shot learning, where the model extracts image patches and clusters them based on their embeddings. To mitigate shortcuts, the few-shot classifier focuses on information that is present within one class but not in other classes rather than relying on potentially spurious correlations present in most of the data. \citet{ma2023eye, ma2023rectify} adapt the attention mechanism in vision transformers to incorporate localization information, enabling the model to focus on regions of the image identified as relevant. \citet{li2023whac} make changes only to the last layer of the model. Instead of relying on a single classification head, they propose to use an ensemble of multiple classifiers trained to mitigate different shortcuts.

\subsubsection{Network Pruning.}
When a model has already been trained on data containing spurious correlations, pruning the network to remove neurons associated with these shortcuts is one approach to mitigate their impact. This approach is more common in the context of backdoor attacks \citep{liu18finepruning, wu21pruning}. Detecting neurons responsible for backdoor triggers can, for example, be done by measuring activations on samples with and without the trigger. 
\citet{linhardt2024preemptively} go even one step further. They prune the model preemptively without specific knowledge about shortcuts in the data. Instead, they use explanation activations to prune the model to focus on important aspects of the input.

\subsubsection{Adversarial Training.} Adversarial training for shortcut mitigation typically involves training a feature extractor alongside a standard classifier and a shortcut predictor. During training, the feature extractor is trained to maximize the accuracy of the standard classifier while minimizing the accuracy of the shortcut predictor \citep{zhao2020training}. In the end, only the base model consisting of the feature extractor and standard classifier is used. \citet{adeli2021representation} introduce an additional loss to minimize the statistical dependence between extracted features and shortcut features. \citet{robinson2021deep} apply adversarial training to mitigate shortcuts in the medical domain where the shortcut predictor tries to differentiate between hospitals as data sources. To mitigate multiple shortcuts at once, \citet{ren2022dice} use a replay buffer in tandem with a shortcut generator to create input samples for the feature extractor and both classifiers. Instead of using an additional shortcut classifier, \citet{minderer2020automatic} use an adversarially trained lens to modify the input to the feature extractor. The lens is trained to minimize the task loss on shortcut data while keeping good reconstruction, assuming that the requirement for good reconstruction asserts that important features are not modified.

\subsubsection{Contrastive Learning.} 
The idea of contrastive (pre-)training is to learn features invariant to shortcuts using standard contrastive learning methods \citep{teney2020learning}. In general, this assumes knowledge about spurious features in the data, and while contrastive learning itself is not task-specific, the information about spurious features implicitly is. \citet{wang2022counterexample} emphasize the role of counterexamples inherently present in the data for contrastive learning. To counteract multimodal shortcuts, \citet{yang2023mitigating} use information about shortcuts in both vision and language simultaneously.

\subsubsection{Explanation-Based Regularization.}
Explanation-based regularization leverages model explanations, like attributions, as proxies to identify which input features the model considers important. Then, the model’s reliance on spurious features is reduced by discouraging high activation of explanations associated with those features. In explanatory interactive learning (XIL), various methods utilize different explanation methods and regularization strategies to mitigate shortcuts \citep{SchramowskiSTBH20, ross2017right, shao2021right} (cf.  \citep{friedrich2023typology} for a detailed overview). \citet{mustafa2024unmasking} use a Jacobian saliency map instead of a binary feedback mask, which is especially beneficial when analyzing brain scans.
\citet{stammer2021right} and \citet{dreyer2024hope} show that regularization based on model explanations is also possible in more high-level, concept-based spaces. While \citet{stammer2021right} use predefined concepts, \citet{dreyer2024hope} employ concept activation vectors as the basis for regluarization.
While the majority of works in this area assume access to full human annotations of the shortcut features, \citet{bassi2024improving} use automated segmentation tools and heuristics to find image backgrounds and penalize the model's explanations to mitigate background-related shortcuts.

\subsubsection{Group Robustness.} Group differential robust optimization (Group DRO) \citep{sagawa2019distributionally} uses group annotations within a dataset to address spurious correlations by identifying samples where target features align or conflict with spurious ones. For example, in the waterbird dataset, groups include combinations like waterbirds in front of water, waterbirds in front of land, landbirds in front of water, and landbirds in front of land. Groups where the shortcut (e.g., waterbirds appearing with water backgrounds) does not apply (such as waterbirds on land) are typically underrepresented and are thus termed minority groups. 
Group DRO introduces an additional loss term that minimizes the empirical worst-group loss to remove the reliance on the shortcut. \citet{zhou2021examining} extends the idea of group DRO beyond a single set of groups. Instead, they optimize over a joint set of groups and features to be more robust against unaccounted shortcuts. To reduce the requirement for extensive group labels, \citet{nam2022spread} uses a small set of group labels to train a classifier that generates pseudo-labels for DRO. \citet{chakraborty2024exmap} go even one step further and circumvent the need for group labels completely by clustering samples with similar explanation heatmaps to obtain pseudo group labels.

\subsubsection{Sample Reweighting.}
Several approaches address shortcuts by training models with weighted samples, aiming to counteract spurious correlations between features originating from under- or overrepresented feature combinations within the dataset. For instance, if waterbirds are more commonly paired with water backgrounds than with land, samples showing waterbirds against a land background are given higher weights to increase their influence during training. Sample weighting can be done with a weighted softmax function \citep{luo2022pseudo}. \citet{kirichenko2022last} and \citet{izmailov2022feature} both use this principle to finetune a model that was previously trained with shortcuts in the data, where finetuning on weighted samples can achieve similar results to finetuning on data without the shortcut \citep{kirichenko2022last}. To obtain information about shortcut or non-shortcut samples, \citet{nam2020learning} and \citet{luo2022pseudo} use a second detector trained via generalized cross-entropy loss, which allows distinguishing between easy-to-learn (shortcut) and harder-to-learn (non-shortcut) samples.

\subsubsection{Feature Reweighting.}
Rather than assigning additional weights to samples, some methods focus on adjusting the weights of specific features during training. The main idea is to down-weight spurious features, encouraging the model to rely on relevant features instead. \citet{chen2022does} utilize information about groups in the data to reweight features according to these groups. \citet{zhang2023robustness} demonstrate that both input-level or high-level features can be reweighted to mitigate shortcuts.  To improve the effectiveness of feature-reweighting, \citet{yang2024identifying} underline the importance of finding groups early in the training and utilizing them continuously during the training process. \citet{asgari2022masktune} perform a very strict form of feature reweighting by removing important features completely during finetuning so that the model explores the use of other features. \citet{holstege2023removing} estimate spurious subspaces of the high-dimensional latent spaces and utilize them to remove spurious concepts.

\subsubsection{Latent-Space splitting.} Several approaches use autoencoder-based models to separate latent representations of spurious features from the remaining information. In these methods, the autoencoder is trained to reconstruct the input while dedicating a part of the latent space to learning spurious features through an additional classifier \citep{yang2022chroma}. As the classifier is regularized to use only a few latent features, it is assumed to rely on shortcuts. The remainder of the latent space is then regularized to encode different information. For the final task, only the non-shortcut part of the latent space is used. Both \citet{yang2022chroma} and \citet{fay2023avoiding} use a mutual information criterion to force the model to encode different information in the shortcut and non-shortcut parts of the latent space. \citet{wang2024navigate} propose to start with a latent space where a part already perfectly encodes shortcuts. During training, the remainder of the latent space then learns non-shortcut features.

\subsubsection{Other Forms of Loss-Based Regularization.} While explanation-based regularization and DRO are the most prominent forms of loss-based regularization to mitigate shortcuts, other approaches have also been explored. After normal training on data with shortcuts, \citet{venkataramani2024causal} finetune only the feature extractor with an additional alignment loss, encouraging similar features on full images and those with only relevant parts visible. With information about a potential counterfactual distribution, \citet{kumar2024causal} estimate the causal effect of features on the output. During training, the model is then regularized to exhibit similar causal effects. Similarly, \citet{tiwari2022robust} regularize the model based on the causal strength between features, such as the impact on one feature when intervening on another. \citet{veitch2021counterfactual} connect shortcut mitigation to counterfactual invariance. They introduce a causal approximation of counterfactual invariance to encourage robustness to small changes in spurious features.

\subsubsection{Miscellaneous.}
\citet{ragonesi2023learning} address shortcuts through meta-learning. In the inner step, they split the data into shortcut-present and shortcut-free sets using pseudo-labeling and train the model to perform well on both. The outer step then combines both sets (via mixup) and updates model parameters on the augmented data. 
\citet{wang2021counterfactual} develop a causality-based training framework to mitigate spurious correlations. They build a structural causal model (given some assumptions like strong ignorability) and enable counterfactual maximum likelihood estimation to ignore shortcuts, utilizing counterfactual information derived from available observations.
\citet{makar2022causally} combine sample weighting and loss regularization. They first derive a distribution without shortcuts from the observational data via sample reweighting and then train the target model with this distribution and causality-motivated loss regularization. 
\citet{arefin2024unsupervised} rely on concepts instead of group labels to mitigate shortcuts. They learn common concepts in the input images in an unsupervised fashion and utilize them instead of the input image to solve the task. This method implicitly assumes that concept learning detects sufficient concepts to solve the task but does not detect shortcut concepts. 
Some works leverage subsets of the data during different training steps. \citet{deng2024robust} start model training on a subset of the data, which is balanced regarding shortcuts. During training, the available data is then progressively expanded so that the model has seen the whole dataset in the end, but without relying on the shortcuts in the dataset.
\citet{moayeri2023spuriosity} sort samples based on shortcut occurrence (given a model trained on the data with these shortcuts). They then finetune the model on the subset of the data with low shortcut occurrence scores. Similarly, \citet{labonte2024towards} construct a subset of the data for finetuning. Instead of relying on explicit confounder information, they train multiple strongly regularized models (e.g., via dropout or early stopping). To select the finetuning set, disagreement between these models is measured.

\subsection{Mitigation at Inference Time}
\label{sec:mitigation_inference}

The third point at which shortcuts can be mitigated is at inference time. While these methods do not come with additional training or dataset preparation costs, they potentially have to be performed every time a model is used for inference. Test-time modifications of, for example, concept-bottleneck models \citep{koh2020concept, chauhan2023interactive} are one example for this category. By intervening on the internal concept representation of these models, shortcuts can be mitigated \citep{steinmann2023learning}.
For large-language models (LLMs), adapting the prompt at inference time can successfully mitigate shortcuts \citep{sun2024exploring}. Additionally, it is possible to utilize retrieval-based approaches to provide additional context to LLMs via an extended prompt. \citet{friedrich2023revision} use this idea to mitigate shortcuts and biases related to moral norms.

Further, in the context of backdoor attacks, augmenting input data with random noise and performing a majority vote on the noisy sample variations has been proposed as a way to receive robust predictions for poisoned samples~\citep{sarkar2020suppression}.

\subsection{Open Challenges in Shortcut Mitigation}
This overview of shortcut mitigation methods reveals a substantial body of work on the topic.
However, as these methods are often developed under different terminologies, they are frequently created in isolation from one another. Consequently, methods are rarely compared against similar and relevant approaches. Additionally, many methods rely on slightly varying assumptions regarding the available data, the information about shortcuts, and the nature of the shortcuts themselves. These differences make it challenging to directly compare methods and assess the overall progress of the field. Comprehensive comparisons with existing methods are therefore a critical step for better structuring and advancing the field.

The typical way to evaluate shortcut mitigation methods is to measure test accuracy on data without the shortcut. In some cases, model reliance on shortcut features is also compared before and after mitigation. However, this evaluation approach has a significant limitation: it primarily confirms whether the model avoids specific known shortcuts but does not assess whether the model is using meaningful and task-relevant features for prediction. Removing identified shortcuts can inadvertently lead models to exploit other, potentially unknown shortcuts \citep{ravichander2023and,li2023whac}. While evaluating whether models rely on relevant features is inherently more challenging, it is a crucial aspect of robust evaluation.

Most shortcut mitigation efforts are concentrated on standard setups where a model is either fully trained or partially trained before undergoing fine-tuning for mitigation. With the widespread adoption of pretraining on large, general-purpose datasets followed by fine-tuning on smaller, task-specific datasets, it remains unclear how shortcuts can be effectively mitigated in these settings. While some studies suggest that general-purpose pretraining can improve model robustness \citep{ghosal2024vision}, other findings suggest that this robustness may not persist through the fine-tuning process \citep{steed2022upstream}. Further research is needed to explore the occurrence and mitigation of shortcuts in the pretraining-fine-tuning setup.

\section{Datasets}
\label{sec:datasets}

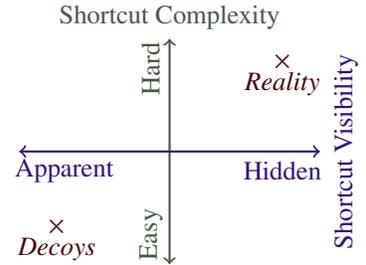
\begin{wrapfigure}{r}{0.35\linewidth}
    \vspace{-0.55cm}
    \definecolor{tradeoffs_color_shortcut}{rgb}{0.27, 0.35, 0.27}
    \definecolor{tradeoffs_color_analysis}{rgb}{0.2, 0.07, 0.48}
    \definecolor{tradeoffs_color_instances}{rgb}{0.28, 0.02, 0.03}
    \begin{center}
        \begin{tikzpicture}
            \draw[<->, thick, color=tradeoffs_color_shortcut] (0,-1.5) -- (0,1.5) node[above] {Shortcut Complexity};
            \draw[<->, thick, color=tradeoffs_color_analysis] (-2,0) -- (2,0) node[right, anchor=center, rotate=90, yshift=-0.35cm] {Shortcut Visibility};
            
            \node[rotate=90, color=tradeoffs_color_shortcut] at (-0.25, 1.1) {Hard};
            \node[rotate=90, color=tradeoffs_color_shortcut] at (-0.25, -1.1) {Easy};
            
            \node[color=tradeoffs_color_analysis] at (-1.4, -0.25) {Apparent};
            \node[color=tradeoffs_color_analysis] at (1.5, -0.25) {Hidden};
            
            \node[font=\itshape, color=tradeoffs_color_instances] at (1.5,1.2) {$\times$};
            \node[font=\itshape, color=tradeoffs_color_instances] at (1.5,0.9) {Reality};
            \node[font=\itshape, color=tradeoffs_color_instances] at (-1.5,-1.0) {$\times$};
            \node[font=\itshape, color=tradeoffs_color_instances] at (-1.5,-1.3) {Decoys};
        \end{tikzpicture}
    \end{center}
    \caption{Dataset Selection Tradeoffs}
    \label{fig:dataset_selection_tradeoffs}
\end{wrapfigure}

In the previous sections, we provided a comprehensive overview of shortcut detection and mitigation methods. To validate the effectiveness of these methods in practice, it is necessary to test them on different datasets. Unlike standard machine learning datasets, those designed for evaluating shortcut learning in a controlled setting must include detailed information about the shortcuts present. To achieve this, some datasets are specifically curated to contain spurious correlations, while others document shortcuts already present in existing ones. Further, datasets differ to which degree contained spurious correlations are identifiable for humans, which is crucial for measuring whether models rely on the wrong reasons \citep{idahl2021towards}.
As \autoref{fig:dataset_selection_tradeoffs} shows, one must often trade off this shortcut visibility to humans with the complexity of the underlying spurious correlations~\citep{shrestha2022investigation}, which can affect how difficult they are to mitigate. To further help dataset selection, we categorize the strength of the shortcut into one of three characteristics most applicable to the prevalent classification tasks:
\emph{perfect}, where a shortcut occurs in only a single class and in all such samples; \emph{semi-perfect}, where a shortcut is also only ever present in one class, but not necessarily always; and \emph{soft}, where possibly $c(f_i, f_j) < 1$ even within a single class. However, it is important to recognize that the difficulty of mitigating a shortcut depends not solely on the dataset but equally on the method.
For instance, while Group DRO can never recover a model confounded by decoys that are present in all samples, that would be a particularly easy case for XIL.

\subsection{Overview of Existing Datasets}

We will cover three groups of publicly accessible datasets.
Firstly, we present widely used general machine learning datasets that have, in hindsight, been found to contain spurious correlations or biases and can now serve as real-world benchmarks (Sec.~\ref{sec:datasets:in_existing}) \citep{lynch2023spawrious}.
However, we do not specifically list these here as they are less suited for controlled studies of their impact on machine learning models.
Instead, we refer readers to more specialized works \citep{quinonero2022datasetShift,koh2021wildsBenchmark,sagawa2022extendingWildsBenchmark,kull2014patterns,ma2024robust}.
While existing datasets with spurious confounders are important case studies, using them for developing or evaluating new methods is challenging.
Secondly, it is highly beneficial for model and mitigation method development to have access to datasets where spurious correlations are easy to spot by humans while simultaneously being favored to be exploited by a machine learning model.
To this end, many popular datasets have been equipped with decoys that synthetically add spurious correlations between, for instance, image patches in specific colors to target classes~\citep{idahl2021towards}.
Thirdly, we cover datasets that still have been specifically assembled to feature spurious correlations. However, the complexity of these correlations often surpasses these of the decoy datasets, with a tradeoff of less human-visibility of the shortcuts.

\subsubsection{Uncovered in Existing Datasets.}
\label{sec:datasets:in_existing}
Reliability is key when deploying machine learning models in medical applications, often motivating a particularly thorough evaluation of learned systems.
This can uncover flaws in existing datasets, such as in various approaches for radiographic COVID-19 detection from lung imagery~\citep{ahmed2021covid19,DegraveJL21}.
Furthermore, various spurious correlations have long plagued skin cancer identification~\citep{reimers2020determining,yan2023trustable}, such as in the case of the ISIC collections~\citep{codella2019skin}, the HAM10000 dataset~\citep{tschandl2018ham10000}, or the ConfDerm dataset~\citep{yan2023trustable}.
These unwanted correlations stem from imperfect data collection procedures, namely different imaging devices in possibly different hospitals, demographic disparities between target and control groups, varying pre-processing steps, additional tags being added for only some of the target classes, and more.
Automated detection of brain tumors using magnetic resonance imaging faced similar challenges, such as the Brain Tumor Dataset~\citep{cheng2015tumor,wallis2022clever}.

Similarly, detrimental spurious correlations have also been found in standard machine learning benchmarks.
For instance, object classification models trained on ImageNet~\citep{russakovsky2015imagenet} have been found to be confounded by the object's texture, whereas humans typically rely on their shape~\citep{geirhos2022imagenet}.
In CelebA~\citep{liu2015celeba}, gender correlates spuriously with hair color \citep{sagawa2019distributionally,izmailov2022feature} and high cheekbones with age and gender \citep{kim2024improving}.
Natural language datasets often contain shortcuts either due to the inherent structure of language or data collection processes.
For instance, the language understanding benchmark MultiNLI~\citep{williams2018multinli} exhibits a strong spurious correlation between the categorization into entailment vs. contradiction and the presence of a negation~\citep{sagawa2019distributionally}.

\begin{table}[tp]
    \centering
    \caption{An overview of commonly employed datasets with explicit spurious correlations.
    The modalities are \underline{Vis}ion, natural \underline{Lang}uage, joint \underline{Vis}ion-\underline{Lang}uage, \underline{Vid}eos, \underline{Log}ic, \underline{Time} Series, and \underline{H}yperspectal \underline{Vis}ion.
    The tasks are \underline{Class}ification, \underline{Lang}uage Modeling, Image \underline{Gen}eration, and \underline{Reason}ing.}
    \label{tab:datasets}
    \small
    \begin{tabular}{l@{\hspace{1pt}}llccccr}
        \toprule
                         & & Name                                          & Modality  & Task   & Characteristic & Origin            & Size \\ 
        \midrule
        \multirow{9}{*}{\rotatebox{90}{\textit{Decoy Datasets}}} & \multirow{9}{*}{\rotatebox{90}{\textit{(Sec.~\ref{sec:datasets:decoy})}}} & ToyColor~\cite{ross2017right} & Vis       & Class  & perfect        & Sampling          & 60k  \\
                         & & Decoy MNIST~\cite{ross2017right}              & Vis       & Class  & perfect        & Sampling          & 70k  \\
                         & & Decoy FMNIST~\cite{teso2019explanatory}       & Vis       & Class  & perfect        & Sampling          & 70k  \\
                        & & CW Decoy FMNIST~\cite{hagos2022impact}        & Vis       & Class  & perfect        & Sampling          & 70k  \\
                        & & Colored MNIST~\cite{arjovsky2019invariant}    & Vis       & Class  & soft           & Sampling          & 70k  \\
                        & & Biased MNIST~\cite{shrestha2022investigation} & Vis       & Class  & soft           & Sampling          & 70k  \\
                        & & Multi-color MNIST~\cite{li2022discover}       & Vis       & Class  & soft           & Sampling          & 70k  \\
                        & & ImageNet-W~\cite{li2023whac}                  & Vis       & Class  & soft           & Sampling          & dyn. \\
                        & & ICD~\cite{liu2024implicitconcept}             & VisLang   & Gen    & perfect        & Sampling          & 1.2M \\
        \midrule
        \multirow{20}{*}{\rotatebox{90}{\textit{Specialized Datasets}}} & \multirow{20}{*}{\rotatebox{90}{\textit{(Sec.~\ref{sec:datasets:specialized})}}} &
          MultiCelebA~\cite{kim2024improving} &
          Vis &
          Class &
          perfect &
          World \& Sampling &
          67k \\
                         & & CelebA hair color~\cite{izmailov2022feature}  & Vis       & Class  & perfect        & World \& Sampling & 183k \\
                         & & CLEVR-Hans~\cite{stammer2021right}            & Vis       & Class  & semi-perfect   & Sampling          & 45k  \\
                         & & ConCon~\cite{busch2024concon}                 & Vis       & Class  & semi-perfect   & Sampling          & 63k  \\
                         & & Spawrious~\cite{lynch2023spawrious}           & Vis       & Class  & semi-perfect   & Sampling          & 152k \\
                         & & NICO~\cite{he2019noniid}                      & Vis       & Class  & soft           & World \& Sampling & 25k  \\
                         & & MetaShift~\cite{liang2022metashift}           & Vis       & Class  & soft           & World \& Sampling & 13k  \\
        & &
          CUB5\textsubscript{box} / CUB5\textsubscript{nat}~\cite{bontempelli2023conceptlevel} &
          Vis &
          Class &
          soft &
          World \& Sampling &
          185k \\
                         & & Waterbirds~\cite{sagawa2019distributionally}  & Vis       & Class  & soft           & World             & 24k  \\
                         & & UrbanCars~\cite{li2023whac}                   & Vis       & Class  & soft           & World             & 8.0k \\
                         & & MetaCoCo~\cite{zhang2024metacoco}             & Vis       & Class  & soft           & Sampling          & 176k \\
                         & & Plant Disease Det.~\cite{SchramowskiSTBH20}   & Vis, HVis & Class  & perfect        & Sampling          & 2.4k \\
                         & & rsbench~\cite{bortolotti2024rsbench}          & Vis, Log  & Reason & perfect        & World \& Sampling & var. \\
                         & & GQA-OOD~\cite{kervadec2021roses}              & Lang      & Class  & soft           & World             & 54k  \\
                         & & CGDialog~\cite{feng2023less}                  & Lang      & Lang   & soft           & World \& Sampling & 0.9k \\
                         & & VQA-VS~\cite{si2022VQA_VS}                    & VisLang   & Class  & soft           & World \& Sampling & 219k \\
                         & & NExT-OOD~\cite{Zhang2024NExTOOD}              & VisLang   & Class  & soft           & World             & 14k  \\
                         & & SCUFO~\cite{li2023scuba}                      & Vid       & Class  & perfect        & World             & 17k  \\
                         & & SCUBA~\cite{li2023scuba}                      & Vid       & Class  & soft           & World             & 17k  \\
                         & & P2S~\cite{kraus2024right}                     & Time      & Class  & perfect        & Sampling          & 2,3k \\ 
        \bottomrule
    \end{tabular}
\end{table}

\subsubsection{Decoy Datasets.}
\label{sec:datasets:decoy}
A series of common machine learning benchmarks have been injected with obvious features that are shortcuts to solving the respective tasks. The first half of \autoref{tab:datasets} provides an overview of these decoy datasets. ToyColor is an early dataset where color patches act as shortcuts to the classification task \citep{ross2017right}. Since then, the classic MNIST~\citep{deng2012mnist}, Fashion MNIST~\citep{xiao2017fmnist}, and ImageNet datasets~\citep{russakovsky2015imagenet} have been artificially confounded for model debugging in various ways each~\citep{ross2017right,teso2019explanatory,arjovsky2019invariant,shrestha2022investigation,li2022discover,hagos2022impact}.
In the large generative text-to-image Implicit Concept Dataset~(ICD), QR codes, watermarks, and text patches are systematically paired with certain target outputs to trigger misguided results~\citep{liu2024implicitconcept}.

\subsubsection{Specialized Datasets.}
\label{sec:datasets:specialized}
A number of datasets has been assembled to contain more challenging shortcuts, as listed in the second half of \autoref{tab:datasets}.
In MultiCelebA~\citep{kim2024improving} and CelebA hair color~\citep{izmailov2022feature}, inherent shortcuts in CelebA have been amplified in the train set and eliminated in the test set by resampling.
This has been analogously performed to obtain the MetaShift dataset~\citep{liang2022metashift}.
The synthetic CLEVR-Hans spuriously correlates object attributes such as color or shape with class labels~\citep{stammer2021right}, making it much more challenging than the original CLEVR reasoning dataset~\citep{johnson2016clevr}.
ConCon extends this to a more difficult continual setting with two degrees of severity~\citep{busch2024concon}.
Many datasets, including the web-scraped NICO~\citep{he2019noniid} and MetaCoCo~\citep{zhang2024metacoco}; the stitched CUB5 variants~\citep{bontempelli2023conceptlevel}, Waterbirds~\citep{sagawa2019distributionally}, and UrbanCars~\citep{li2023whac}; as well as the generated Spawrious dataset~\citep{lynch2023spawrious} spuriously correlate target objects with background types or co-occurring objects.
In the real-world biology setting of Plant Disease Detection, the background of common and hyperspectral imagery is a shortcut for determining plant health~\citep{SchramowskiSTBH20}.
In the context of vision-based logic and reasoning, the rsbench suite can be used~\citep{bortolotti2024rsbench}.
The small CGDialog~\citep{feng2023less} and large GQA-OOD~\citep{kervadec2021roses} test sets filter and resample existing textual datasets to reveal confounded models.
This is extended to visual question answering by the datasets VQA-VS~\citep{si2022VQA_VS} and NExT-OOD~\citep{Zhang2024NExTOOD}.
For example, one shortcut is that a common answer to \enquote{How many \dots?} is \enquote{2}.
To assess to which degree video classification models fall for shortcuts, the SCUBA and SCUFO datasets replace the background and freeze the frames while injecting distracting foreground objects, respectively~\citep{li2023scuba}.
Finally, the Production Press Sensor~(P2S) dataset contains manufacturing time series where the quality of the finished good is spuriously correlated with the operation speed of the press~\citep{kraus2024right}.

\subsection{Future Directions}

The presented overview shows that current works provide datasets that predominantly contain pictures, and the vast majority pose classification tasks.
Using images to analyze spurious correlations is certainly appealing, as humans tend to excel at many vision tasks where machines struggle---for instance, at ignoring decoys.
However, modern applications may face shortcuts in many other modalities, such as natural language~\citep{yuan2024llms}, graphs~\citep{fan2023generalizing}, audio recordings \citep{chettri2023clever}, time series\citep{kraus2024right}, and increasingly multimodal combinations thereof \citep{dancette2021beyond,si2022VQA_VS}.
Moreover, they are not confined to the primary data domain; they can also manifest in other forms, such as in the frequency domain \citep{wang2023dfm}.
Similarly, today's tasks are much more diverse than classification, encompassing many more supervised tasks, including regression, summarization, ranking, and forecasting; self-supervised tasks, such as language modeling and generative modeling; and unsupervised tasks, such as clustering and density estimation.
To effectively develop methods for mitigation and detection in these other settings, collecting datasets is an important prerequisite.
However, obtaining high-quality data with explicit annotations about shortcuts is highly resource-intensive \citep{lee2024clarify}.
The already high resource demands of data collection are amplified by the challenges of identifying unknown or subtle shortcuts for labeling purposes \citep{adebayo2022post}.
This might explain why many of the existing datasets are of only limited size.

Furthermore, current datasets make it extremely difficult to learn grounded correlations and decide for the right reasons. One useful perspective for this is that causal mechanisms are often deeply hidden in the data and hard to extract without further knowledge. However, models can only learn relevant correlations when the datasets contain enough information to reveal these patterns from purely observational data. This can be achieved, e.g., by featuring non-deterministic effects~\citep{lippe2023biscuit}, excluding any unobserved confounders in the causal sense~\citep{lippe2023biscuit}, or sufficient variability~\citep{zhang2024causalrep}.
Considering such properties when collecting datasets would offer a more principled approach to evaluating models for shortcut learning.

Many studies use the few existing datasets in slightly differing ways, with varying assumptions about available annotations and different levels of spurious correlations.
This inconsistency makes it notoriously difficult to compare methods fairly and effectively \citep{shrestha2022investigation}.
Moreover, comparisons between methods in comparable settings are rarely conducted, further complicating the evaluation process.
Establishing unified evaluation protocols would significantly strengthen the rigor of developing shortcut mitigation and detection methods.

\section{Outlook and Open Challenges}\label{sec:outlook}

In this section, we outline key directions for future research in this field. Beyond the specific opportunities discussed in the detection, mitigation, and dataset sections, we highlight broader challenges and potential directions to advance the field.

\paragraph{Beyond Image Classification.}
As reviewed in earlier sections, most work on shortcut detection and mitigation has focused on vision data and classification tasks. However, shortcuts can occur across various learning settings, including reinforcement learning \citep{ding2024seeing,langosco2022misgeneralization,delfosse2024interpretable,delfosse2024hackatari}, out-of-distribution detection \citep{zhang2021deep}, and sequential or continual learning scenarios, where spurious correlations can be even harder to identify and address \citep{busch2024concon}. Furthermore, shortcuts are not confined to the vision domain; they also arise in other areas, such as natural language processing and time-series data. It is, therefore, critical to expand the scope of shortcut research to encompass diverse tasks, domains, and modalities.

\paragraph{Shortcuts in Generative Models.} 
The increasing adoption of generative models like large language models (e.g., GPT4 \citep{openai2024gpt4technicalreport}, Llama3 \citep{touvron2023llama}) and diffusion models (e.g., stable diffusion \citep{rombach2022high}) underscores the importance of understanding their reaction to shortcuts. Initial studies indicate that LLMs can also exhibit shortcut behavior in various settings \citep{yuan2024llms,yuan-etal-2024-llms,Du2024shortcut}. Similarly, text-to-image diffusion models are prone to biases in training data, such as generating stereotypical outputs (e.g., associating poverty exclusively with dark skin tones, even when queried otherwise \citep{bianchi2023easily,seshadri-etal-2024-bias}).
The increasing reliance on massive web-scraped datasets for model pretraining \citep{schuhmann2022laion} raises the likelihood of both biases \citep{birhane2024into, friedrich2023fair} and shortcuts \citep{dogra2024shortcut} being included in the data. Addressing these shortcuts in generative models presents unique challenges, as traditional dataset- and training-focused mitigation techniques are not directly applicable to large-scale pre-trained models. Consequently, expanding shortcut detection and mitigation to generative models and large datasets is an essential step.

\paragraph{More Complex Shortcuts.} 
Most existing methods for detecting and mitigating shortcuts focus on similar shortcuts: spurious correlations between an input and target feature. However, not all shortcuts follow the same pattern, and their complexity can significantly impact method applicability. For example, group robustness approaches often assume the existence of a worst-case group (samples without the shortcut) \citep{sagawa2019distributionally}. These methods fail when confronted with perfect shortcuts, i.e., shortcuts that are present in all samples. Moreover, many existing techniques are designed to address a single shortcut, while real-world datasets frequently contain multiple co-occurring shortcuts. Methods that focus solely on mitigating individual shortcuts may prove insufficient in such cases \citep{li2023whac}. Future research should aim to develop and evaluate methods capable of handling these more complex shortcut scenarios.

\paragraph{Moving Towards Improved Task Specifications.}
As we have outlined, one primary reason shortcuts arise is that machine learning tasks are often imprecisely defined, serving as proxies rather than directly specifying the intended goals. This lack of precision leaves models vulnerable to learning unintended cues, as there are no explicit guidelines to prevent such behavior. A potential step to mitigate the emergence of shortcuts is to formulate tasks more precisely. While this is challenging, particularly when working with raw input data, adopting neuro-symbolic approaches focusing on object-centric and conceptual representations can help simplify the process. Task formulation \citep{wustpix2code}, as well as shortcut detection \citep{anders2022finding} and mitigation \citep{stammer2021right, stammer2024neural}, can become more manageable within such conceptual frameworks. However, even symbolic representations are not immune to reasoning shortcuts \citep{marconato2024not}, highlighting the ongoing need for rigorous validation. Overall, efforts to improve shortcut detection and mitigation, as well as reducing opportunities for models to learn shortcuts, are both valid ways to address the problem.

\section{Conclusion}\label{sec:conclusion}
In this work, we presented a comprehensive overview of shortcut learning. By introducing formal definitions of \textit{shortcuts} and their underlying \textit{spurious correlations}, we clarified these terms and established clear connections to related concepts such as \textit{confounders} and \textit{Clever Hans} behavior. To structure this fragmented and often confusing field, we introduced a unified and comprehensive taxonomy of shortcut learning, organizing approaches from various sub-areas into a coherent structure.

Our taxonomy not only organizes the current body of research but also highlights critical gaps and challenges in the field. Notably, studying more complex shortcuts, expanding evaluations beyond classification tasks, and developing methods applicable to domains outside of vision are essential for progress. By collecting and analyzing datasets with explicit information about the shortcuts they contain, we aim to equip researchers with valuable resources to develop and evaluate new approaches.

In summary, this work serves as a foundation for unifying and advancing the field of shortcut learning, connecting theoretical insights with practical approaches. By addressing the outlined challenges and leveraging the opportunities presented, the machine learning community can move towards systems that do not rely on shortcuts but make decisions for the right reasons instead.